\documentclass[a4paper, 10pt, conference]{ieeeconf}      

\IEEEoverridecommandlockouts                              
\usepackage{graphicx}
\usepackage{amsmath} 
\usepackage{amssymb}  
\usepackage[utf8]{inputenc}
\usepackage{balance}
\usepackage[detect-all]{siunitx}

\usepackage{times}
\usepackage{epsfig}
\usepackage{booktabs}
\usepackage{placeins}

\newcommand{\etal}{\textit{et al}. }

\title{\LARGE \bf
DFuseNet: Deep Fusion of RGB and Sparse Depth Information for Image Guided Dense Depth Completion
}

\author{Shreyas S. Shivakumar, Ty Nguyen, Ian D. Miller, Steven W. Chen, Vijay Kumar and Camillo J. Taylor
\thanks{We gratefully acknowledge the support of Novateur Research Solutions Sub to Air Force Contract FA9550-18-C-0050}
\thanks{Shreyas S. Shivakumar, Ty Nguyen, Ian D. Miller, Steven W. Chen, Vijay Kumar and Camillo J. Taylor are with the GRASP Laboratory, School of Engineering and Applied Sciences,
        University of Pennsylvania, Philadelphia PA 19104
        {\tt\small \{sshreyas,tynguyen,iandm,chenste,kumar,cjtaylor\}
        @seas.upenn.edu}}%
}

\begin{document}

\maketitle
\thispagestyle{empty}
\pagestyle{empty}

\begin{abstract}
In this paper we propose a convolutional neural network that is designed to upsample a series of sparse range measurements based on the contextual cues gleaned from a high resolution intensity image. Our approach draws inspiration from related work on super-resolution and in-painting. We propose a novel architecture that seeks to pull contextual cues separately from the intensity image and the depth features and then fuse them later in the network. We argue that this approach effectively exploits the relationship between the two modalities and produces accurate results while respecting salient image structures. We present experimental results to demonstrate that our approach is comparable with state of the art methods and generalizes well across multiple datasets.
\end{abstract}

\section{INTRODUCTION}

Dense depth estimation is a critical component in autonomous driving, robot navigation and augmented reality. Popular sensing schemes in these domains involve a high resolution camera and a low resolution depth sensor such as a LiDAR or Time-of-Flight sensor. The density of points returned from commonly available depth sensors is typically an order of magnitude lower than the resolution of the camera image. Additionally, higher resolution variants of these sensors are expensive, making them impractical for most applications. However, a number of applications such as planning and obstacle avoidance can benefit from higher resolution range data which motivates us to consider approaches that can up-sample the sparse available depth measurements to the resolution of the available imagery.

Traditionally, interpolation and diffusion based schemes have been used to up-sample sparse points into a smooth dense depth image, often using the corresponding color image as a guide~\cite{ferstl2013image}. Convolutional neural networks (CNN) have had tremendous success in depth estimation tasks using monocular image data~\cite{godard2017unsupervised,zhang2018deep,li2018monocular,li2018deep,li2018megadepth,fu2018deep}, stereo image data~\cite{song2018edgestereo,chang2018pyramid,zbontar2016stereo,kendall2017end} and sparse depth data on it's own~\cite{uhrig2017sparsity,mal2018sparse,ku2018defense,eldesokey2018propagating}.


One way to view both the monocular depth prediction problem and the depth completion problem is in terms of a posterior distribution $P(D|I)$ which represents the probability of a given depth image, $D$, given an input intensity image, $I$. 
In both cases the approaches implicitly assume that the resulting posterior distribution is highly concentrated along a low dimensional manifold which makes it possible to infer the complete depth map from relatively few depth samples

We wish to design a CNN architecture that can learn sufficient global and contextual information from the color images and use this information along with sparse depth input to accurately predict depth estimates for the entire image, while enforcing edge preservation and smoothness constraints. Once designed, such a network could be used to upsample information from a variety of depth sensors including LiDAR systems, stereo algorithms or structure from motion algorithms. To summarize, we propose the following contributions:
\begin{enumerate}
    \item A CNN architecture that uses a dual branch architecture, spatial pyramid pooling layers and a sequence of multi-scale deconvolutions to effectively exploit contextual cues from the input color image and the available depth measurements.
    \item A training regime that can make of use different sources of information, such as stereo imagery, to learn how to extrapolate depth effectively in regions where no depth measurements are available.
    \item An evaluation of our methods on the KITTI Depth Completion Benchmark\footnote{http://www.cvlibs.net/datasets/kitti/eval\_depth.php},
    virtual KITTI, NYUDepth, and our own mini-dataset. We show that our method is able to generalize well across these four different datasets.
    \item We also make our mini-dataset publicly available as it may serve as an additional source of validation data to the community. Details can be found in: \emph{https://github.com/ShreyasSkandanS/DFuseNet}.
\end{enumerate}

\begin{figure*}
\begin{center}
   \includegraphics[width=0.90\linewidth]{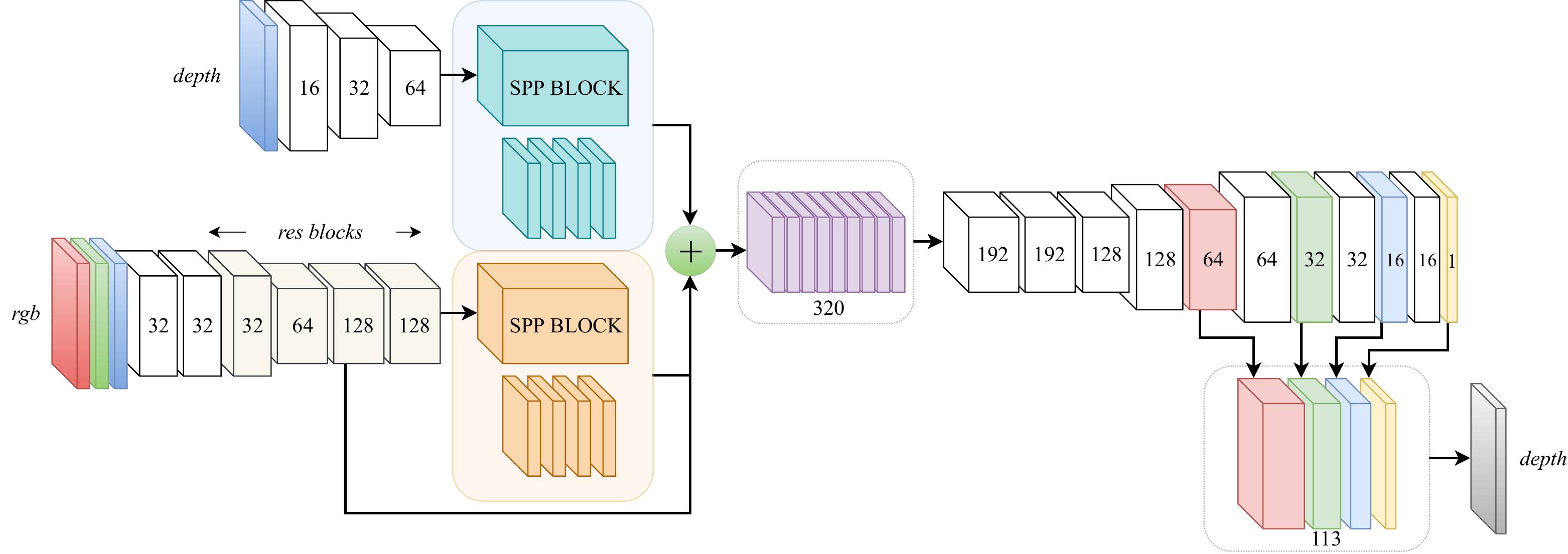}
\end{center}
   \caption{Our network architecture uses two input branches for RGB depth input respectively. We use Spatial Pyramid Pooling (SPP) blocks in the encoder and use a hierarchical representation of decoder features to predict dense depth images.}
\label{fig:network_architecture}
\end{figure*}
\section{RELATED WORK}

\textit{Depth Estimation:}
Monocular depth estimation is an active research field where CNN based methods are currently the state of the art. Different methods have been proposed that use supervised ~\cite{eigen2014depth,liu2016learning, chen2016single,kong2018pixel,li2018monocular,zhang2018deep}, unsupervised ~\cite{godard2017unsupervised} and self-supervised~\cite{li2018megadepth} depth estimation strategies. At time of writing, the best performing monocular depth estimation algorithm is from Fu \etal, achieving an inverse RMSE score of 12.98 on the KITTI depth prediction dataset~\cite{fu2018deep}. 

CNNs have been successfully used in dense stereo depth estimation tasks. Zbontar \etal proposed a siamese network architecture to learn a similarity measure between two input patches. This similarity measure is then used as a matching cost input for a traditional stereo pipeline~\cite{zbontar2016stereo}.
Recently, many end-to-end methods have been proposed that are able to generate accurate disparity images while preserving edges ~\cite{kendall2017end,song2018edgestereo,yang2018segstereo,khamis2018stereonet}. The work of Chang \etal is similar to the network we propose, where the authors propose an end-to-end approach using pyramid pooling to better learn global image dependent features~\cite{chang2018pyramid}.

\textit{Incomplete Input Data:}
Learning dense representations from sparse input is similar to the domain of super resolution and in-painting. Super resolution assumes that the input is a uniformly sub-sampled representation of the desired high resolution output, and the learning problem can be posed as an edge preserving interpolation strategy. A comprehensive review of these methods is presented by Yang \etal \cite{yang2014single}. We note that multi-scale architectures with multiple skip connections have been successfully used for image and depth upsampling tasks \cite{yamanaka2017fast,hui2016depth}.
Content-aware completion is motivated by a similar problem of learning complete representations from incomplete input data. Image in-painting requires semantically aware completion of missing input regions. Generative networks have been used successfully for context aware image completion tasks~\cite{yeh2017semantic,yu2018generative} but are outside the scope of this paper. 

\textit{Depth Completion:} A particular sub-problem of depth estimation with incomplete input data is depth completion. Following the release of the KITTI depth completion benchmark, novel approaches to solve the problem have been proposed. Uhrig \etal \cite{uhrig2017sparsity}, the authors of the benchmark, propose a sparsity invariant CNN architecture, using partial normalized convolutions on the input sparse depth image. They propose multiple architectures that accommodate RGB information and sparse depth input only. Huang \etal propose HMSNet, which uses masked operations on the partial convolutions such as partial summation, up-sampling and concatenation~\cite{huang2018hms}.


Ku \etal propose a non-learning based approach to this problem to highlight the effectiveness of well crafted classical methods, using only commonly available morphological operations to produce dense depth information \cite{ku2018defense}. Their proposed method currently out-performs multiple deep learning based methods on the KITTI depth completion benchmark. Dimitrievski \etal propose a CNN architecture which uses the work of Ku \etal as a pre-processing step on the sparse depth input \cite{dimitrievski2018learning}. We followed a similar strategy and chose to fill in our sparse input depth image instead of using sparse convolutions. Their network is designed to use traditional morphological operators as well as subsequently learned morphological filters using a U-Net style architecture \cite{ronneberger2015u}. They are able to achieve better quantitative results but their model fails to preserve semantic and depth discontinuities as it relies heavily on the filled depth image for their final output. Eldesokey \etal propose a method that also uses normalized masked convolutions, but generates confidence values for each predicted depth by using a continuous confidence mask instead of a binary mask \cite{eldesokey2018propagating}. A similar confidence mask based approach is proposed by Gansbeke \etal \cite{van2019sparse}. Cheng \etal propose a depth propagation network to explicitly learn an affinity function and apply it to the depth completion problem~\cite{cheng2018depth}. 

Wang \etal propose a multi-scale feature fusion method for depth completion~\cite{wang2018multi} using sparse LIDAR data. Ma \etal propose two methods, a supervised method for depth completion using a ResNet based architecture~\cite{mal2018sparse} and a self-supervised method~\cite{ma2018self} which uses the sparse LiDAR input along with pose estimates to add additional training information based on depth and photometric losses.
\section{Approach}

\subsection{Design Overview}

We propose the CNN architecture depicted in Figure ~\ref{fig:network_architecture}.  It has been structured to learn local to global context information from both the color image and the sparse depth data as well as to fuse them together to produce accurate and consistent dense depth maps. We propose a dual branch encoder design in a similar fashion to previous image comparison networks ~\cite{zbontar2016stereo}. Given the differences in input modality provided to the two branches, we choose to not use Siamese networks with coupled weights~\cite{bromley1994signature}, and instead use independent branches with different design decisions made for each branch. In our encoder, we use spatial pyramid pooling (SPP) blocks to learn a coarse-to-fine representation of features. Spatial pyramid pooling blocks have been effective in learning local to global context information and have been successfully used in depth perception tasks~\cite{chang2018pyramid}. We concatenate features learned from individual branches and propagate these features through our de-convolution layers. The final layer performs a convolution operation on features combined from different de-convolution layers, up-sampled to the final output resolution, to utilize information from different scales and context to generate the final depth image.
\subsection{Feature Extraction}


Our color and depth branches begin with an initial depth filling step, similar to the approach of Ku \etal ~\cite{ku2018defense}. We use a simple sequence of morphological operations and Gaussian blurring operations to fill the holes in the sparse depth image with depth values from nearby valid points such that no holes remain. This is then passed to the feature extraction branch. The filled depth image is then normalized by the maximum depth value in the dataset, resulting in depth values between 0 and 1. For the depth image, we choose to use larger kernel sizes and fewer convolution operations, resulting in fewer layers. For the color image, we use smaller kernel sizes and make use of four residual blocks~\cite{he2016deep}, in addition to two initial convolution layers. The output of these initial feature extraction layers is then passed to spatial pyramid pooling (SPP) blocks. We use a similar structure to that proposed by Chang \etal \cite{chang2018pyramid}, but use max pooling for our depth branch and average pooling for our color branch.
Our pooling windows are consistent between the two branches and are 64, 32, 16 and 8 for each scale respectively. The output of this layer is an up-sampled stack of feature layers carrying information from different scales. 

\begin{figure}[t]
\begin{center}
\includegraphics[width=0.47\linewidth]{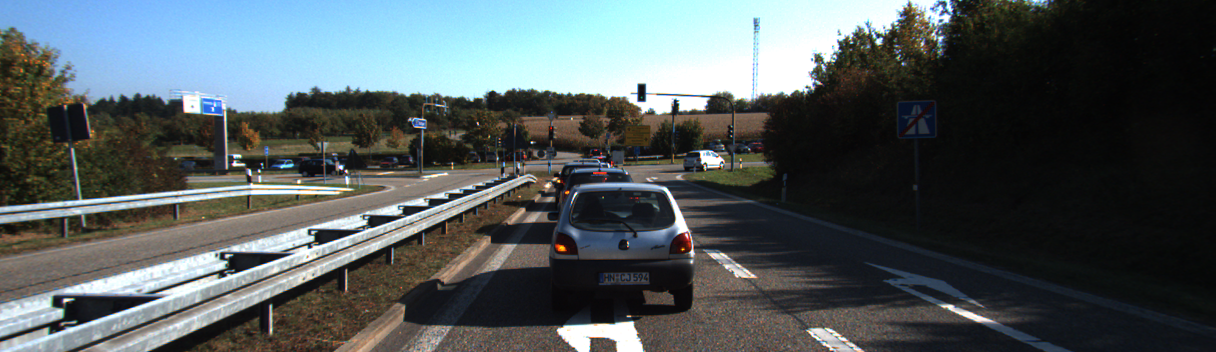}
\includegraphics[width=0.47\linewidth]{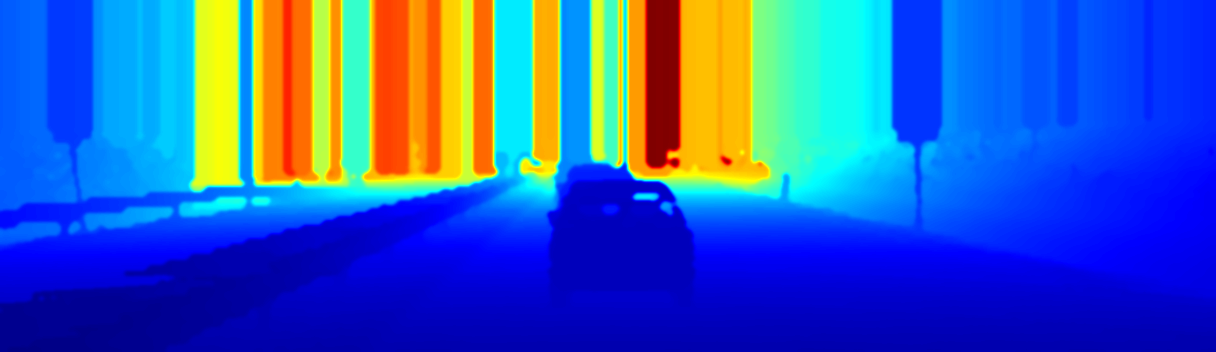}
\includegraphics[width=0.47\linewidth]{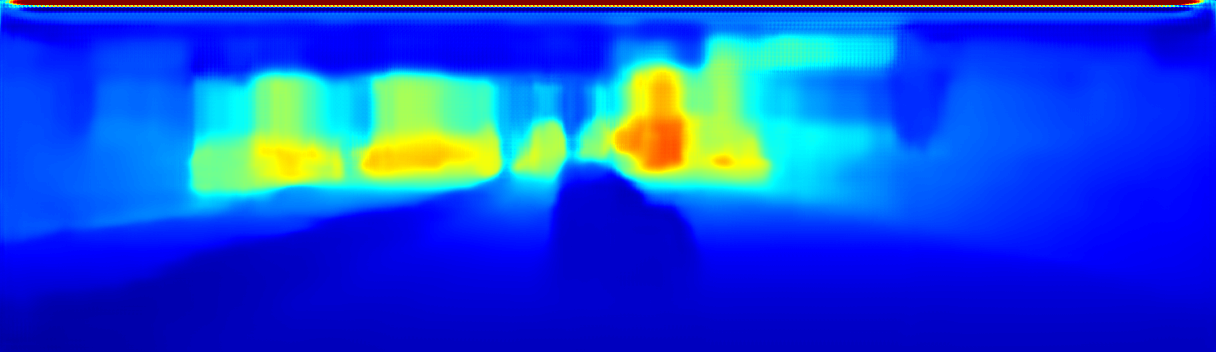}
\includegraphics[width=0.47\linewidth]{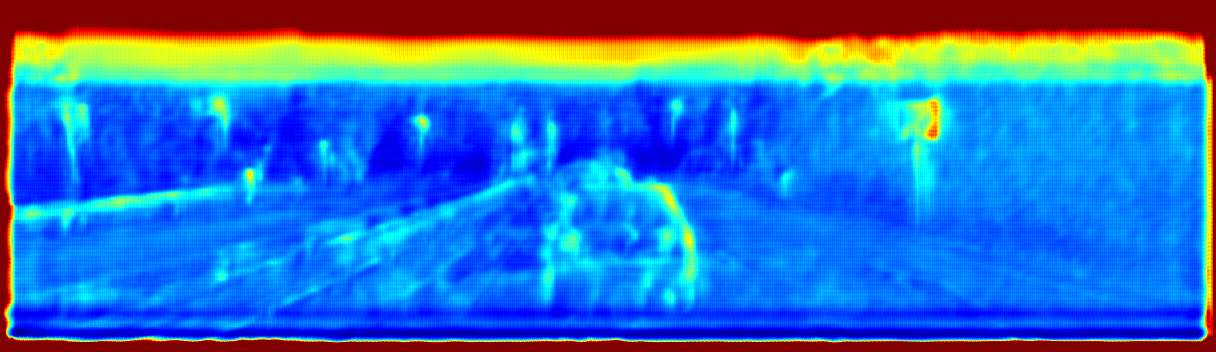}
\includegraphics[width=0.95\linewidth]{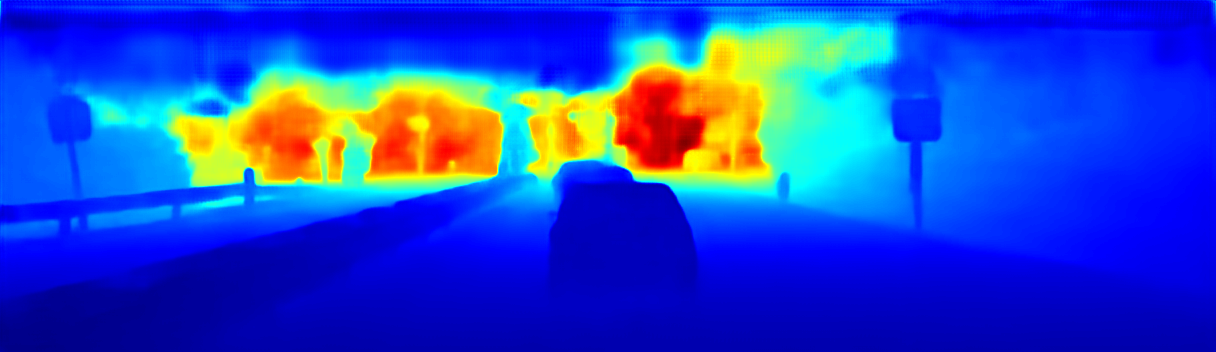}
\end{center}
   \caption{Dual-branch architecture (L-R,T-B): Input color image, filled input depth, output when input to RGB branch is set to zeros, output when input to the depth branch is set to zeros and predicted depth image when RGB and depth images are provided. This illustration informs us that both branches contribute significantly to the final prediction and that the filled depth is not being naively propagated through the network without any learning.}
\label{fig:dual_branch}
\end{figure}

\subsection{Combining Modalities}

The features from the previous extraction modules are then concatenated into one volume. The first layer is an intermediate output of the residual blocks from the color branch, which we hypothesize can carry over high level features learned from the color image. The subsequent layers are color and depth features extracted from the SPP blocks of the two branches. We believe that these layers can help learn a joint feature representation between the two input modalities in the following layers. We perform three sequential convolution operations on this volume, reducing the number of channels and increasing the spatial resolution by twice the size of the volume. By forcing a reduction in channels we attempt to force the network to learn a lower dimensional representation of the joint feature space, combining important information from both depth and color branches.

\subsection{Depth Prediction}

The following layers perform a sequence of convolutions with batch normalization, and incremental de-convolutions to restore the original image resolution. The final step involves concatenating different layer outputs from the de-convolution pipeline, up-sampling by interpolation to achieve the original input resolution and then performing a final convolution on the multi-scale stack to produce a single channel output. This output is then passed to a sigmoid activation function and re-scaled to the original range of depth values. Odena \etal advise caution in the use of transposed convolutions for spatial upsampling~\cite{odena2016deconvolution}, hence we limit the use of transposed convolutions and our final output is a result of a 1x1 convolution on a feature volume, which mainly consists of interpolated low resolution features, and hence minimizes the checkerboard effect in the final depth image. 

\subsection{Training}

Our training signal $\mathcal{L}_{train}$ is a weighted average of multiple loss terms, some calculated over the entire image resolution and some calculated only at points where accurate ground truth depth exists. The weights $\alpha$, $\beta$ and $\gamma$ are chosen based on a confidence associated with each signal and are varied at different points in time in the training regime.
\begin{equation}
    \mathcal{L}_{train} = \alpha\mathcal{L}_{primary} + \beta\mathcal{L}_{stereo} + \gamma\mathcal{L}_{smooth}
\label{eq:main_loss}
\end{equation}

\subsubsection{Primary Loss}

We experimented with both L1 and L2 norms as primary loss functions $\mathcal{L}_{primary}$. For this term, we calculate the loss only at pixels where ground truth depth exists and average over the total number of ground truth points.  For better RMSE values on evaluation benchmarks we found L2 to be the better choice as a primary loss term. 

\begin{figure}[t]
\begin{center}
\includegraphics[width=0.49\linewidth]{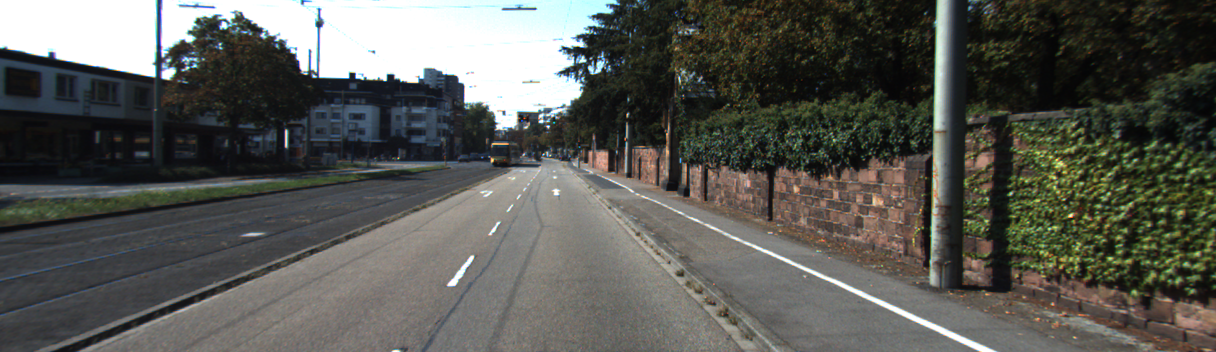}
\includegraphics[width=0.49\linewidth]{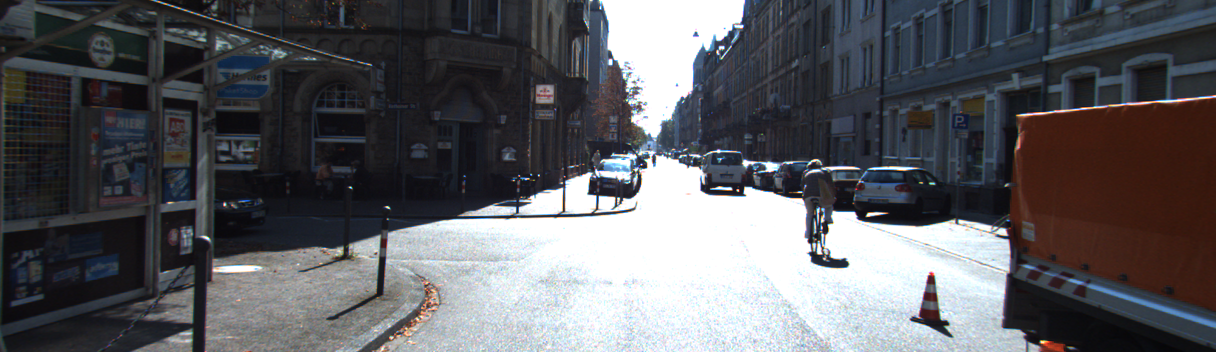}
\includegraphics[width=0.49\linewidth]{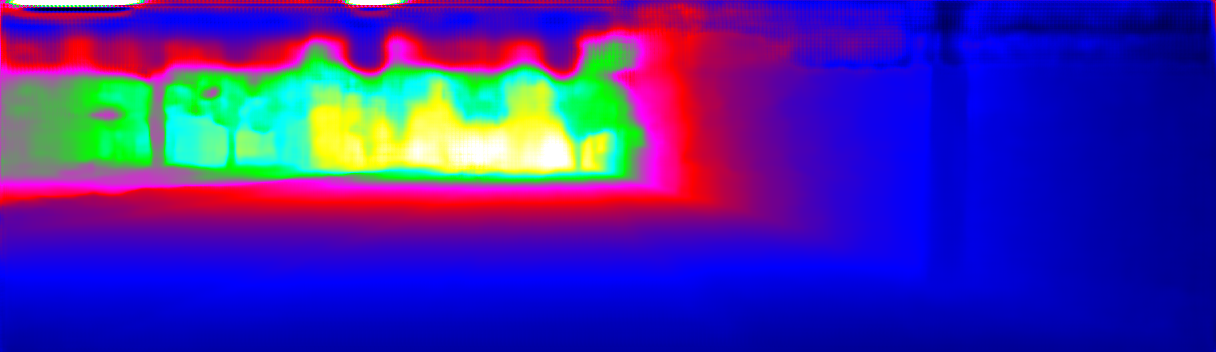}
\includegraphics[width=0.49\linewidth]{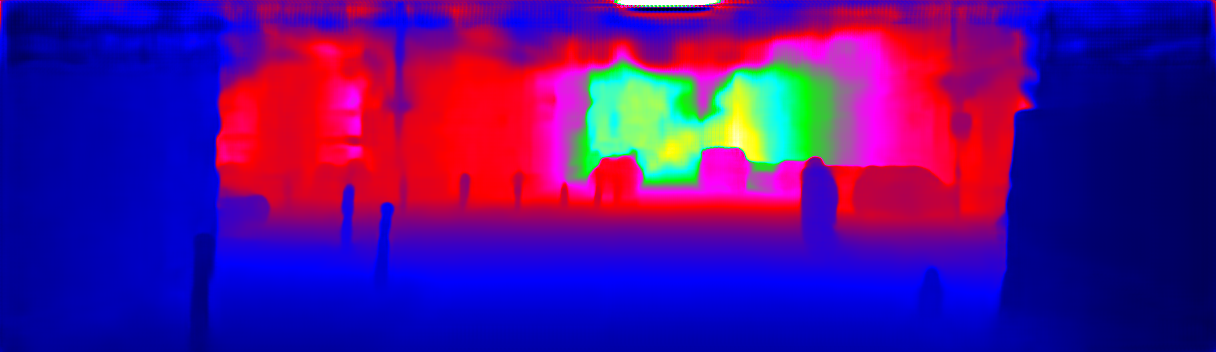}
\includegraphics[width=0.49\linewidth]{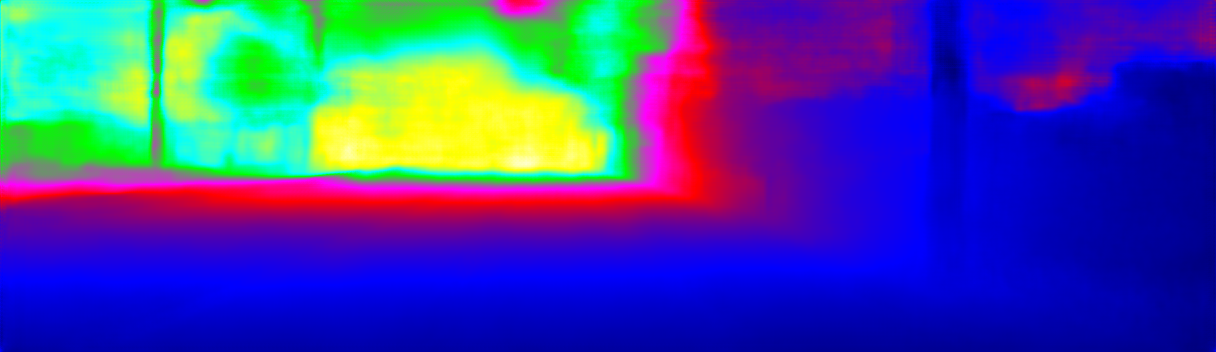}
\includegraphics[width=0.49\linewidth]{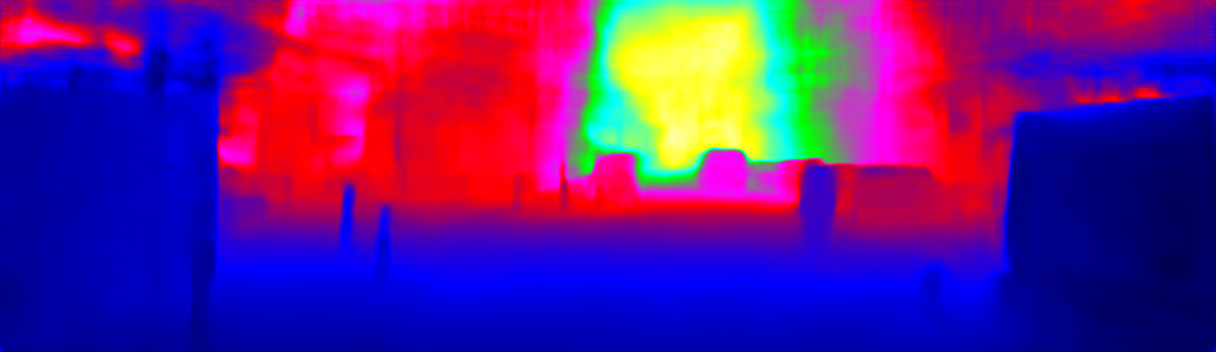}

\end{center}
   \caption{Learning to extrapolate better using available information: By adding a stereo depth based loss term, we are able to make better extrapolations in regions where no ground truth or LiDAR exists. (T-B) Input image, predicted depth without stereo term and prediction with stereo term.}
\label{fig:stereo_extrapolate}
\end{figure}

\subsubsection{Optional Stereo Supervision}

Since Uhrig \etal provide a large dataset with data from multiple cameras, we propose a means of making better use of this data during training. The KITTI depth completion dataset provides roughly 42k stereo image pairs, and we use these images to provide depth information at points where ground truth LiDAR data is missing. 
We propose an auxiliary loss term $\mathcal{L}_{stereo}$ that uses the stereo input image pair to generate a dense depth estimate that can guide the learning process in regions where no ground truth LiDAR measurements exist. We compute this loss term in a self-supervised manner since stereo intrinsics and extrinsics are known. We use Semi Global Matching to generate this dense depth estimate~\cite{hirschmuller2005accurate}. This loss term is an L2 norm of the difference between the predicted depth and the stereo estimated depth. This term can be computed at almost every pixel in the input image. Some pixels lack depth estimates since we use left-right consistency checks to discard noisy and partially occluded depth estimates.

\begin{figure*}
\begin{center}
\includegraphics[width=0.32\linewidth]{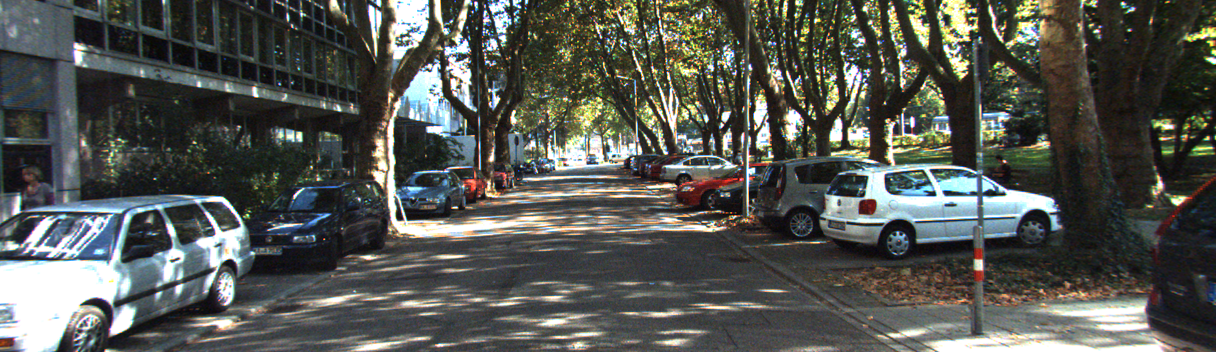}
\includegraphics[width=0.32\linewidth]{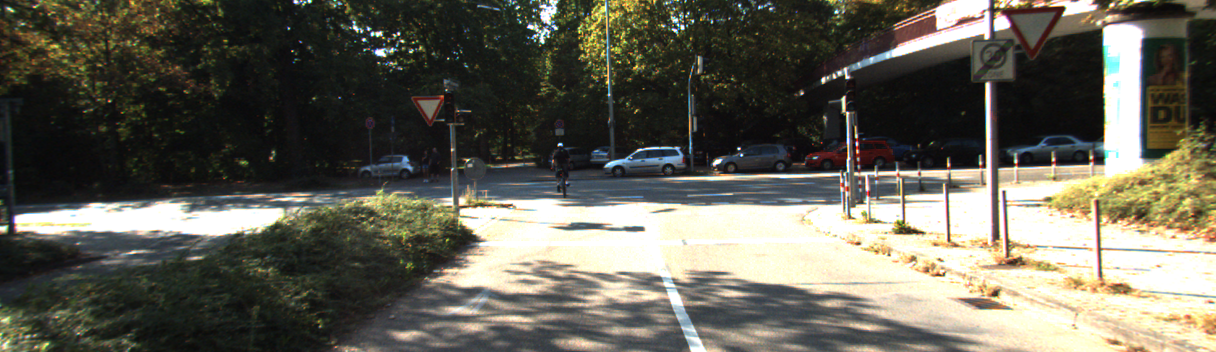}
\includegraphics[width=0.32\linewidth]{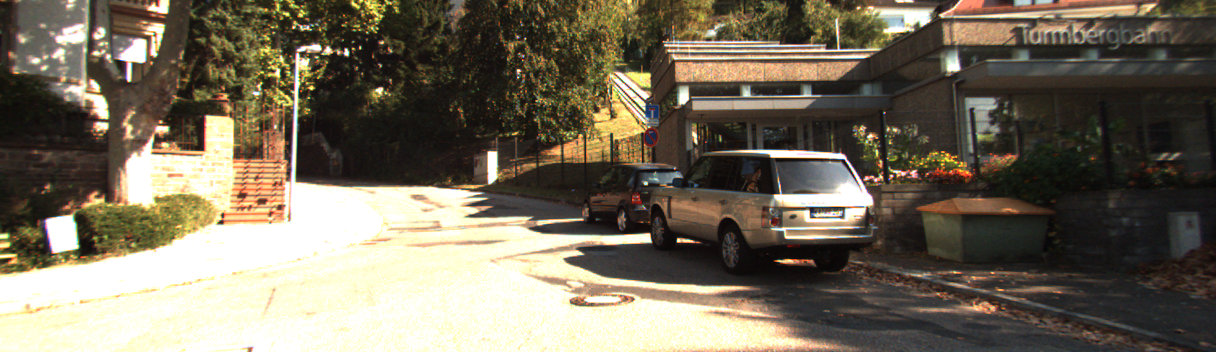}
\includegraphics[width=0.32\linewidth]{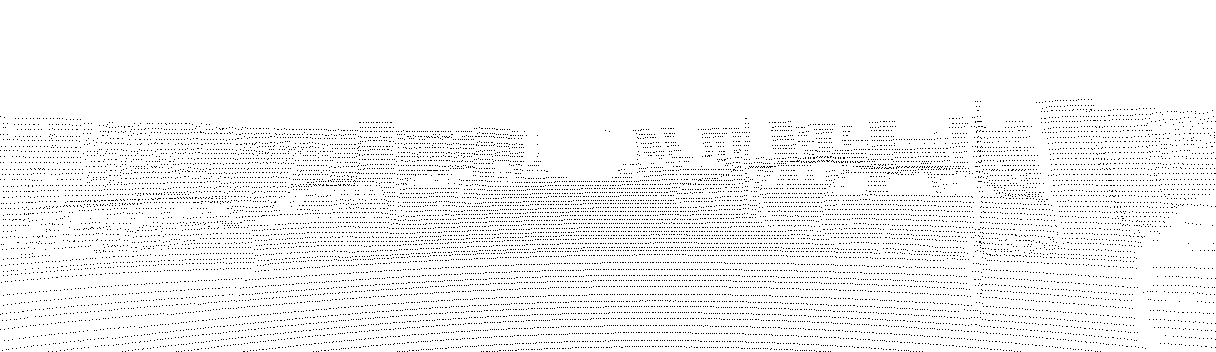}
\includegraphics[width=0.32\linewidth]{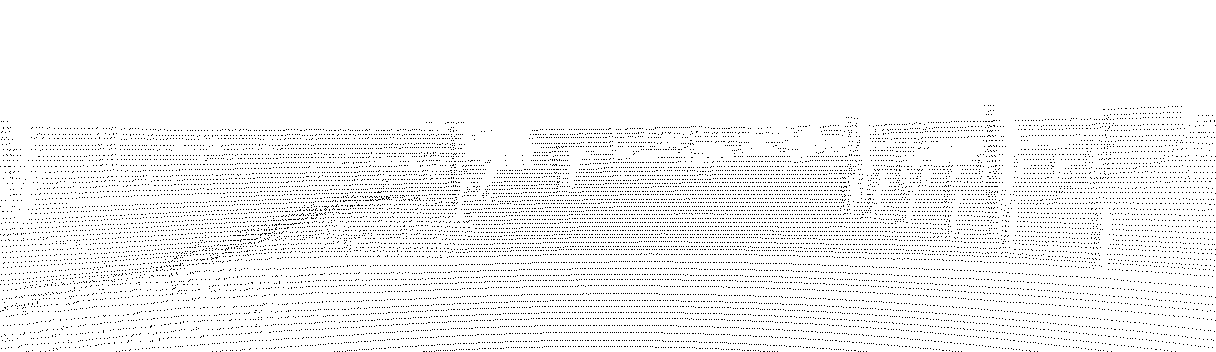}
\includegraphics[width=0.32\linewidth]{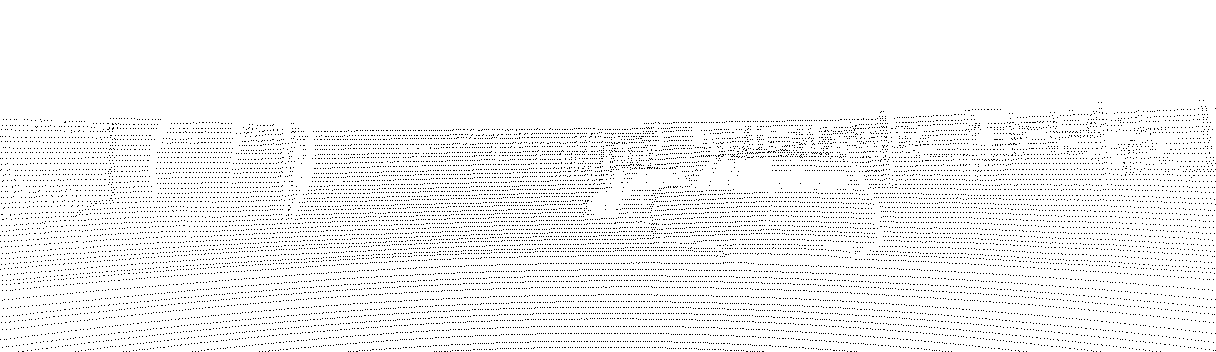}
\includegraphics[width=0.32\linewidth]{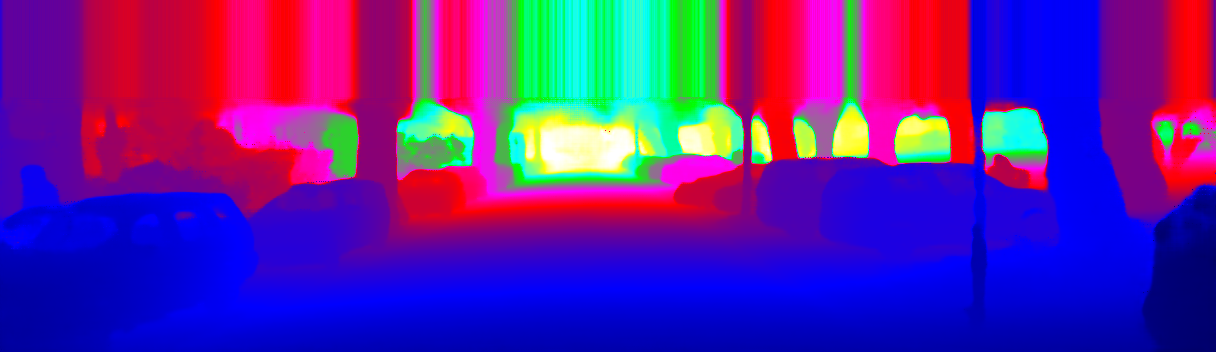}
\includegraphics[width=0.32\linewidth]{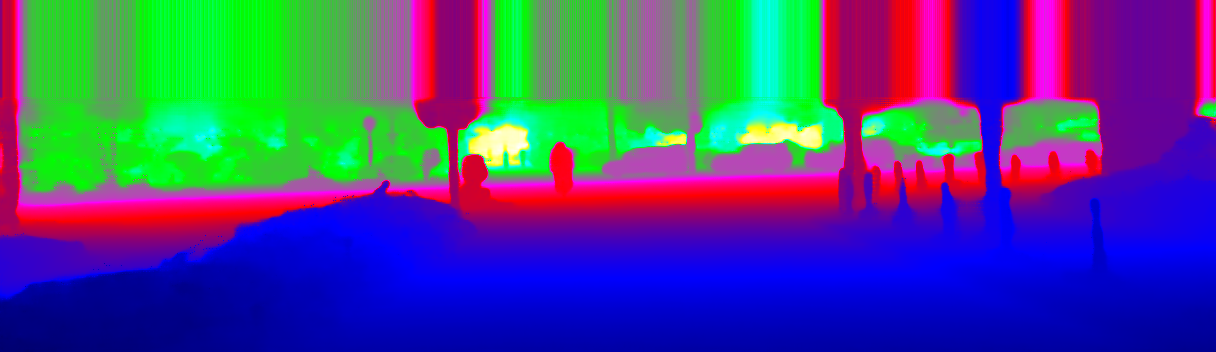}
\includegraphics[width=0.32\linewidth]{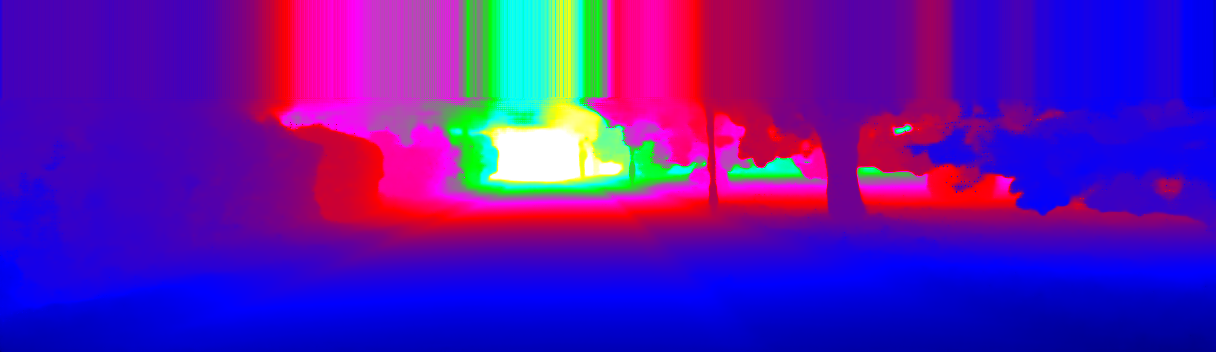}
\includegraphics[width=0.32\linewidth]{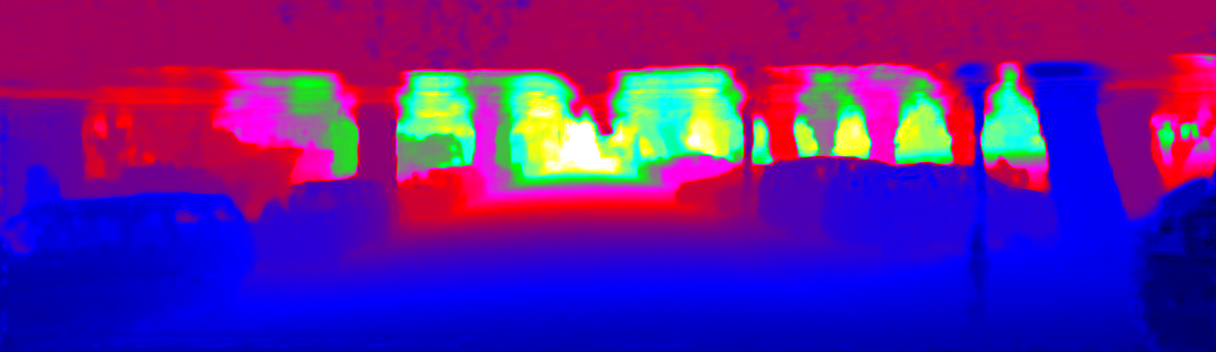}
\includegraphics[width=0.32\linewidth]{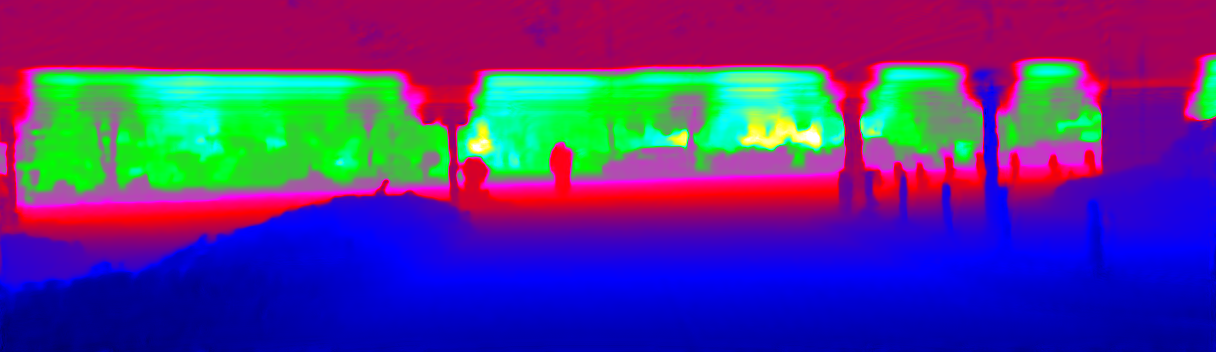}
\includegraphics[width=0.32\linewidth]{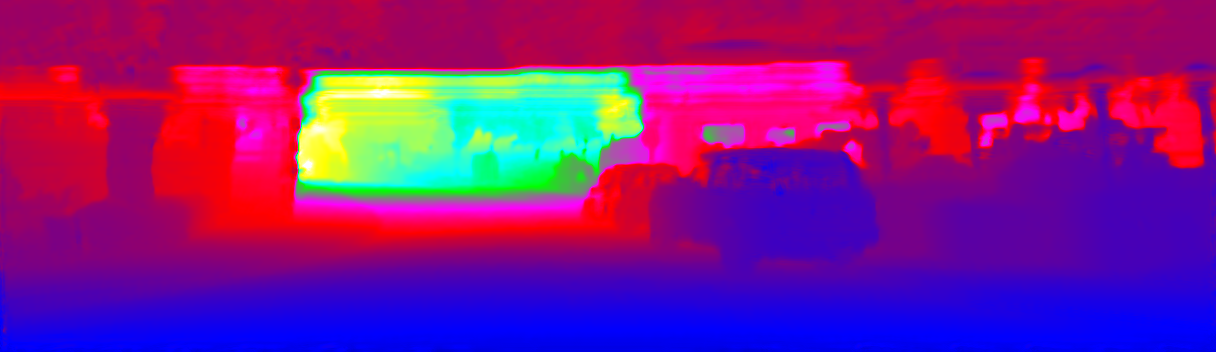}
\includegraphics[width=0.32\linewidth]{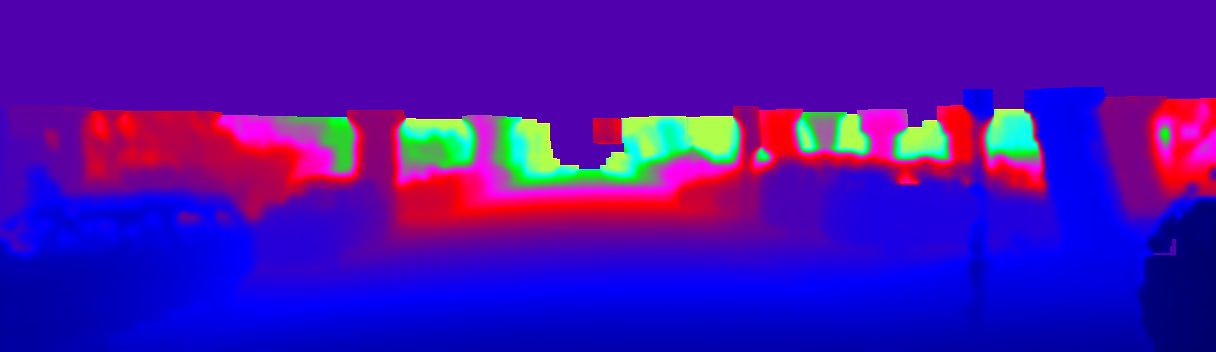}
\includegraphics[width=0.32\linewidth]{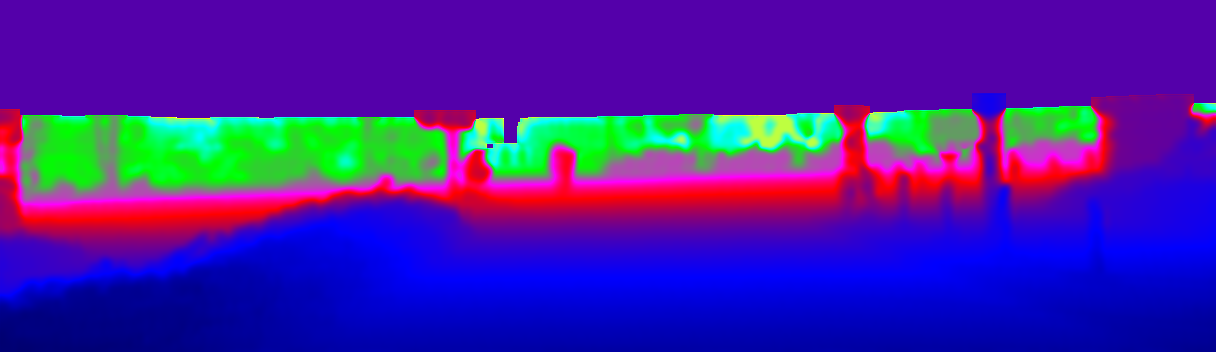}
\includegraphics[width=0.32\linewidth]{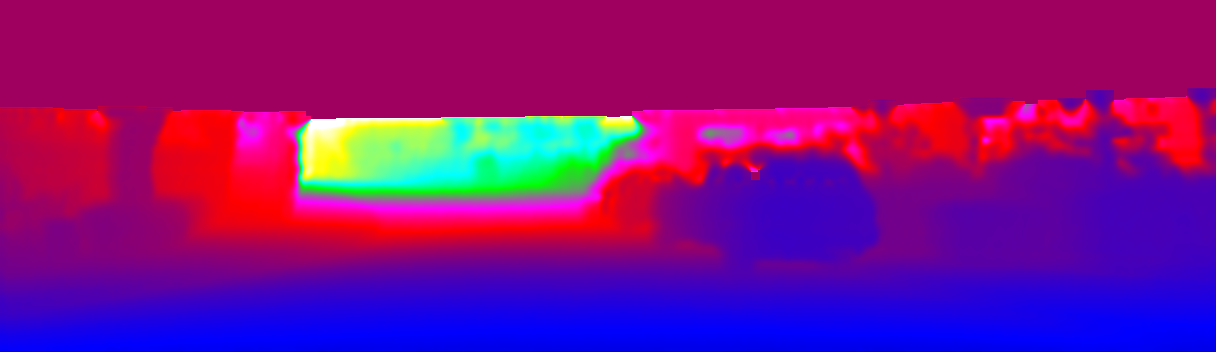}
\includegraphics[width=0.32\linewidth]{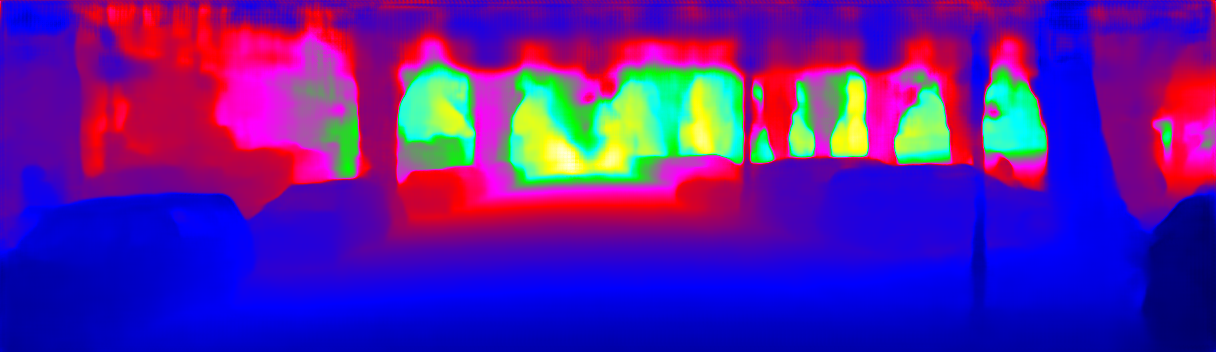}
\includegraphics[width=0.32\linewidth]{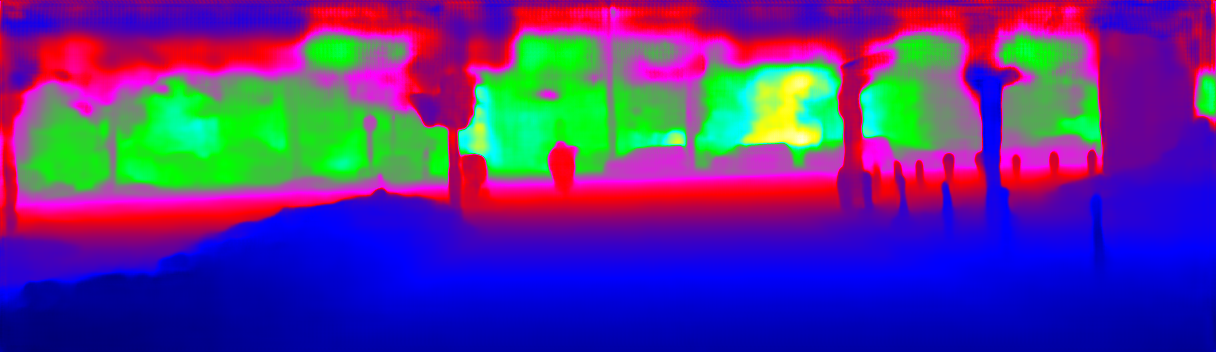}
\includegraphics[width=0.32\linewidth]{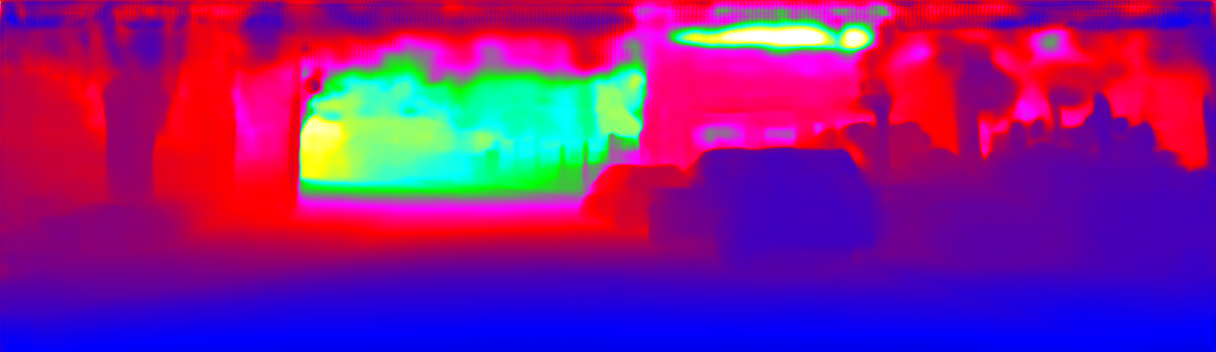}
\end{center}
   \caption{KITTI Depth Completion Results: Top-Bottom: Input color image, corresponding LiDAR scan mask (inverted for visualization), Van Gansbeke \etal ~\cite{van2019sparse}, MorphNet~\cite{dimitrievski2018learning}, SparseConv~\cite{uhrig2017sparsity} and Ours (DFuseNet)}
\label{fig:kitti_comp}
\end{figure*}

\subsubsection{Smoothness}

We add a smoothness loss term $\mathcal{L}_{smooth}$ , which is an L1 norm on the second order derivative of the predicted dense image, similar to the strategy used in unsupervised monocular depth estimation and structure from motion networks ~\cite{godard2017unsupervised, zhou2017unsupervised, vijayanarasimhan2017sfm}.

\section{EXPERIMENTS}

\textit{Implementation Details:}
All our networks were implemented in PyTorch\footnote{http://pytorch.org} and we train them from scratch, not using pre-trained weights for any layers. Our models are trained using the ADAM optimizer, and we typically use batch sizes of 20-25 for our experiments and train for roughly 40 epochs for all experiments. We use an initial learning rate of $10^{-4}$, and drop our learning rate by a factor of 10\% after every 5 epochs. We use a weight decay term of $10^{-4}$. The weight terms from Eq \ref{eq:main_loss} are: $\alpha$ is usually set to 1, $\beta$ is 0.01 and $\gamma$ is 0.001.

\subsection{KITTI Depth Completion}


The ground truth depth provided in the KITTI Depth Completion dataset is created by merging 11 LiDAR scans from frames before and after a given frame using pose estimates provided in the dataset \cite{uhrig2017sparsity}. These projected 3D points are refined using stereo depth estimation algorithms to discard outliers. During evaluation, the final scores are based only on these refined ground truth LiDAR points. 

While this corpus provides a large amount of training data the available range measurements are typically clustered towards
the bottom of the available imagery and are often missing at critical contextual regions such as object boundaries. A consequence is that models trained on this data often produce blurry edges since the available measurements and evaluation tools do not contraindicate such solutions. 

Additionally the data set does not provide information in distant regions like the sky and many previous approaches involve cropping out regions where no LiDAR data is available. In contrast we seek to preserve as much contextual information as possible and make depth predictions across as much of the image as possible using all available data.

\subsubsection{Quantitative Comparison}

Our method achieves a mean RMSE score of \SI{1206.66}{mm} and the current published state of the art is \SI{772.87}{mm}~\cite{van2019sparse}. 
We are quantitatively out performed by Sparse-to-Dense~\cite{ma2018self}, NConv-CNN~\cite{eldesokey2018propagating}, HMS-Net~\cite{huang2018hms}, CSPN~\cite{cheng2018depth} and MorphNet~\cite{dimitrievski2018learning}, but we believe that our model is able to better incorporate RGB image information to generate edge preserving and semantically smooth depth images at the cost of a small loss in metric accuracy. We highlight this in Figure \ref{fig:kitti_comp}, where it is clear that our method is able to use contextual information to preserve semantic boundaries as well as or better than methods that outperform us on the benchmark.
\subsubsection{Learning to extrapolate with limited ground truth data}

We validate the effectiveness of our stereo based loss term by comparing our model with and without this term. Quantitatively the improvements are minimal, i.e the model trains faster and results in slightly improved accuracy, qualitatively we noticed that our network can now extrapolate depth values at regions where no input LiDAR scan or ground truth exist. This is specially useful in datasets such as KITTI where the ground truth information is semi-dense with significant regions of the image missing ground truth LiDAR points. In Figure \ref{fig:stereo_extrapolate} we show a qualitative comparison of our network demonstrating its ability to extrapolate beyond the range of the LIDAR scans.

\subsection{Virtual KITTI}

We evaluate our network on the Virtual KITTI dataset~\cite{gaidon2016virtual}. This dataset contains roughly 21k image and depth frames generated in virtual worlds with simulated lighting and weather conditions, in a driving dataset similar to KITTI . The maximum depth range for this dataset is \SI{655.3}{m} (sky), but for simplicity and similarity to our previous dataset, we set our perception limit to \SI{100}{m} and train our model accordingly. We use 60\% of this data as our training set and evaluate our model on the remaining images. To generate sparse depth input, we randomly sample 10\% of the ground truth depth data uniformly. We apply the same input filling step as in the previous dataset, using the same parameters and morphological window sizes. We then pass this filled depth image along with the RGB image to our network and evaluate our accuracy in the \SI{0}{m}-\SI{100}{m} range.

\begin{figure}
\begin{center}
\includegraphics[width=0.48\linewidth]{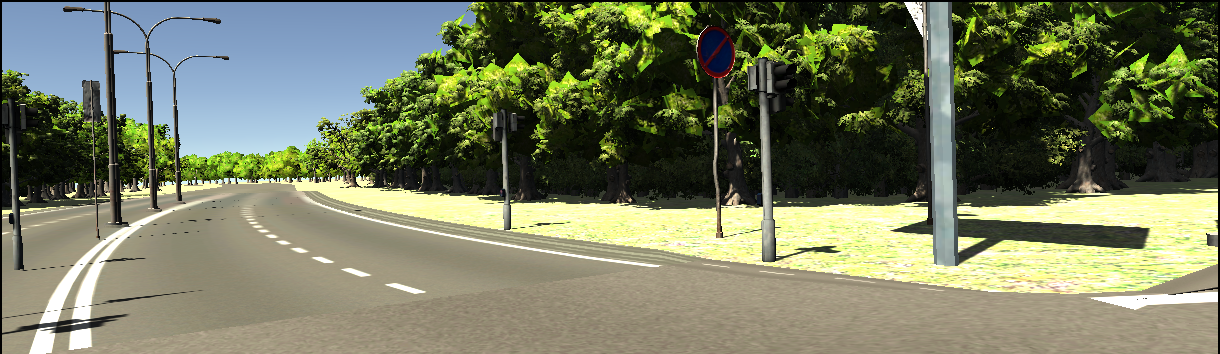}
\includegraphics[width=0.48\linewidth]{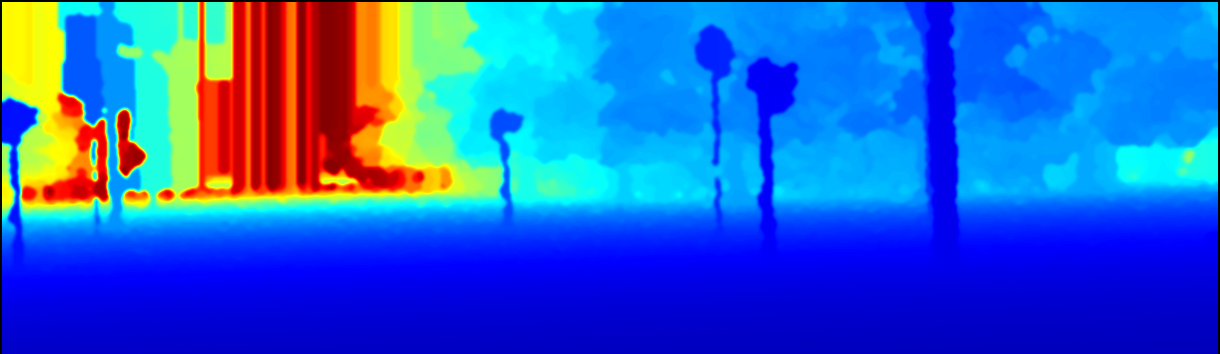}
\includegraphics[width=0.48\linewidth]{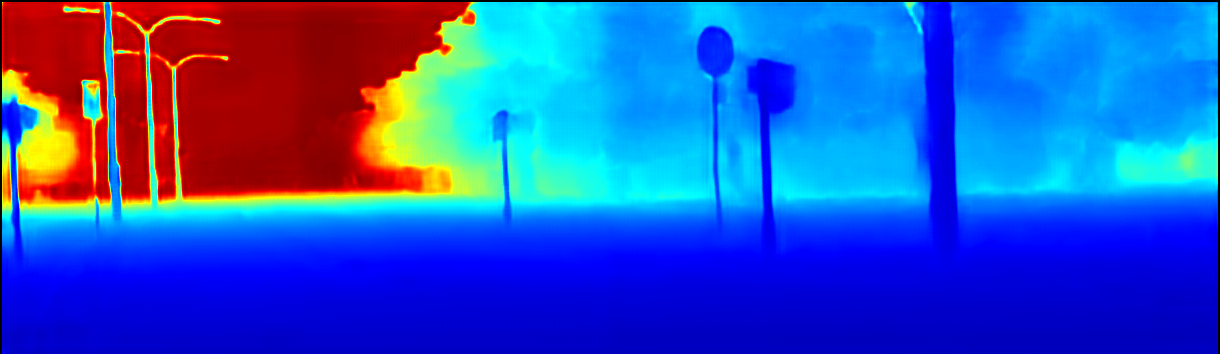}
\includegraphics[width=0.48\linewidth]{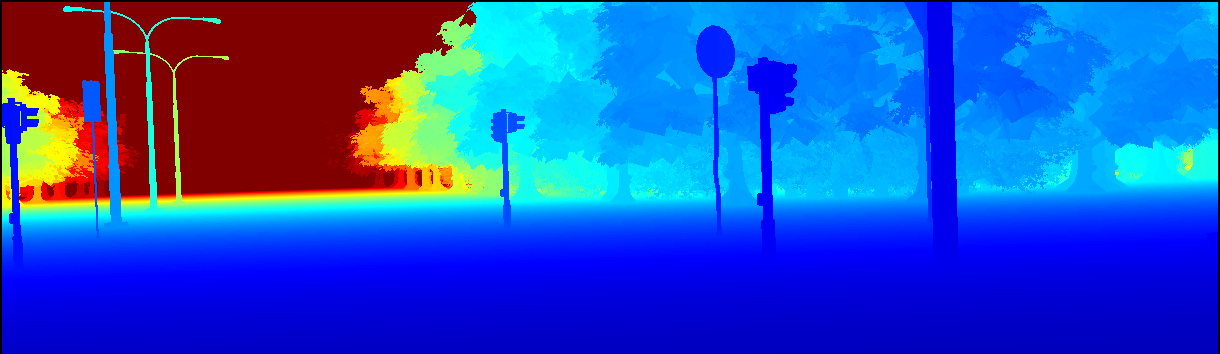}
\end{center}
   \caption{Virtual KITTI Results (L-R,T-B): Color image, input sparse depth (10\% randomly selected ground truth points) image with filled depth after pre-processing, prediction and ground truth}
\label{fig:vkitti_images}
\end{figure}

While the virtual KITTI dataset is not an accurate representation of real life data, we show that our method is able to learn to accurately generate depth dense images, while preserving edges and contextual information. Figure \ref{fig:vkitti_images} shows the our results on this dataset. We achieve an RMSE of \SI{3.458}m and MAE of \SI{1.355}m on our validation set.

\subsection{NYU Depth V2}

In our evaluation on the NYUDepthV2 dataset~\cite{silberman2012indoor}, we use the 1449 densely labelled pairs of aligned RGB and depth images, and split our dataset into 70\% training and 30\% validation. All our errors are reported on the 30\% validation set and we compare our errors against the errors reported by other authors in their respective papers~\cite{cheng2018depth,mal2018sparse,liao2017parse}. We use the full resolution 640$\times$480 images as our input and use the same method of subsampling as above to generate sparse input depth measurements from the ground truth. We use this dataset to verify that our model is able learn in different environments using different sources of input data, since here a Kinect RGBD sensor is used to collect data in various common environments such as offices and homes.

Table \ref{table:nyudepthresults} shows the performance of our model at multiple levels of sparsity compared to the work of Ma \etal and Liao \etal at 200 samples~\cite{mal2018sparse,liao2017parse}. 
Our approach performs comparably to the the approach of Ma \etal and better than that of Liao \etal. 
We use the same morphological window size and operations as in the Virtual KITTI and KITTI datasets and our method is able to generate accurate results even with noisy input filling. Again the filling process helps us by removing all zeros in the depth image and providing a reasonable initialization but the final depth prediction is based on the combined features from the RGB and depth branches of the network. It must be noted here that the results reported here were computed using a different randomly chosen set of samples and a direct comparison would be unfair.

\begin{table}[!t]
\caption{NYUDepthV2 Results: Comparisons are made to the errors reported by respective authors. Note: the authors use different training and validation sample sets, and errors here were not computed on the same data.}
\label{table:nyudepthresults}
\centering
\resizebox{0.475\textwidth}{!}
{
\begin{tabular}{cccccc}
\textbf{Method} - No. of samples  & \textbf{RMSE} & \textbf{REL} & \textbf{$\delta_{1}$} & \textbf{$\delta_{2}$} & \textbf{$\delta_{3}$}\\
\midrule 
DFuseNet (ours) - 200 & 0.2966 & 0.0609 & 0.9588  & 0.9927 & 0.9982\\
DFuseNet - 500 & 0.2195 & 0.0441 & 98.04 & 99.70 & 99.93\\
DFuseNet - 1k & 0.1759 & 0.0371 & 98.78 & 99.82 & 99.96\\
\midrule
Cheng et.al (rgbd)~\cite{cheng2018depth} - 500 & 0.117 & 0.016 & 99.2 & 99.9 & 100.0\\
Ma \etal (rgbd)~\cite{mal2018sparse} - 200 & 0.230 & 0.044 & 97.1 & 99.4 & 99.8\\
Liao \etal (rgbd)~\cite{liao2017parse} - 225 & 0.442 & 0.104 & 87.8 & 96.4 & 98.9\\
\bottomrule
\end{tabular}
}
\end{table}

\subsection{Number of depth samples}

\begin{figure*}
\begin{center}
\includegraphics[width=0.20\linewidth]{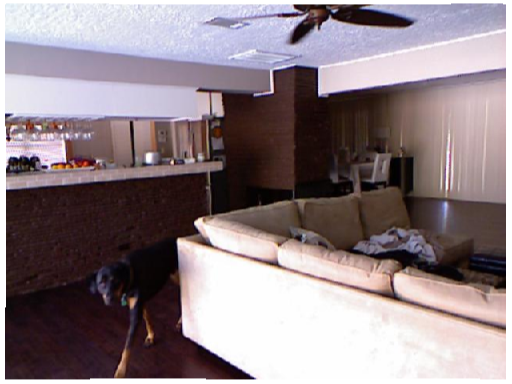}
\includegraphics[width=0.20\linewidth]{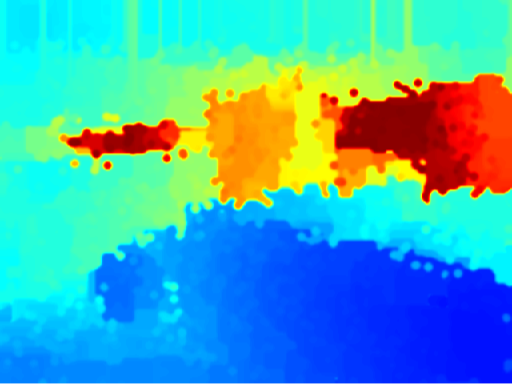}
\includegraphics[width=0.20\linewidth]{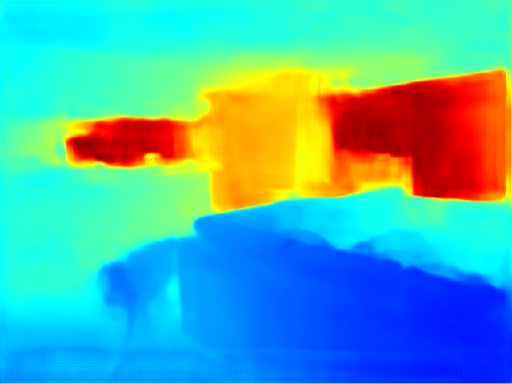}
\includegraphics[width=0.20\linewidth]{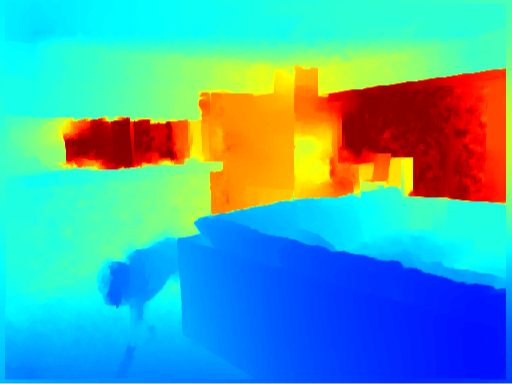}
\includegraphics[width=0.20\linewidth]{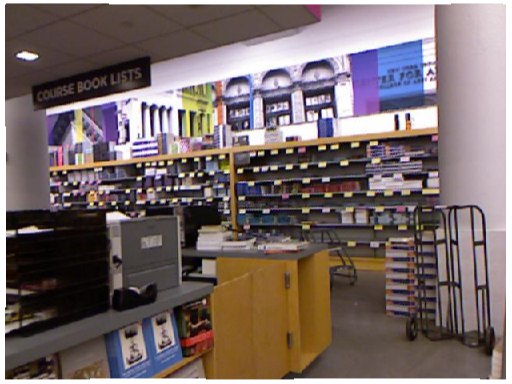}
\includegraphics[width=0.20\linewidth]{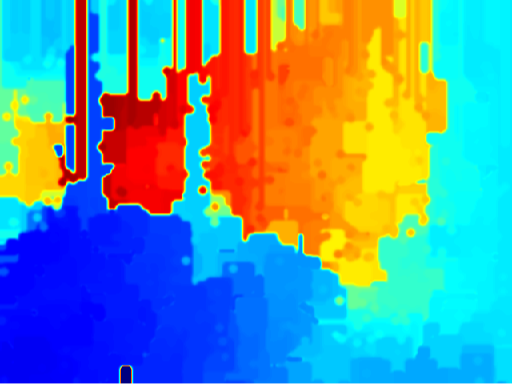}
\includegraphics[width=0.20\linewidth]{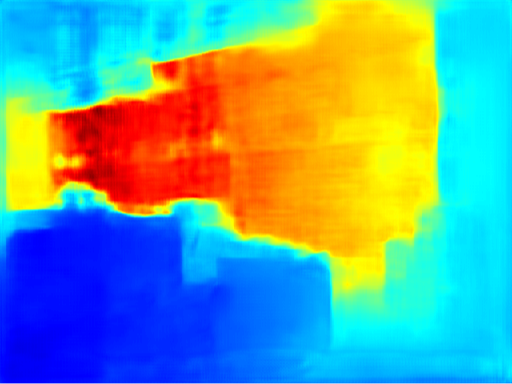}
\includegraphics[width=0.20\linewidth]{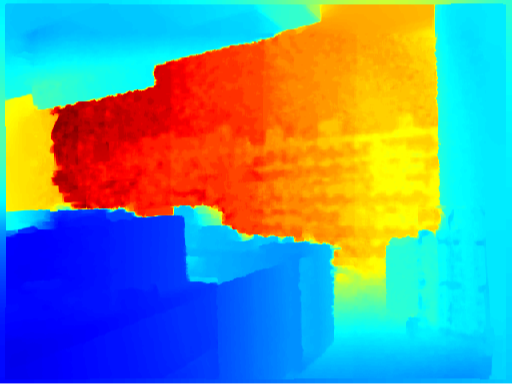}
\end{center}
   \caption{NYUDepthV2 Results with 1000 input depth samples (first row) and 500 input samples (second row). We use the same depth filling parameters as in our previous datasets; L-R: Input image, filled input depth, prediction and ground truth.}
\label{fig:nyudepth_images}
\end{figure*}

\begin{figure}[t]
\begin{center}
\includegraphics[width=\linewidth]{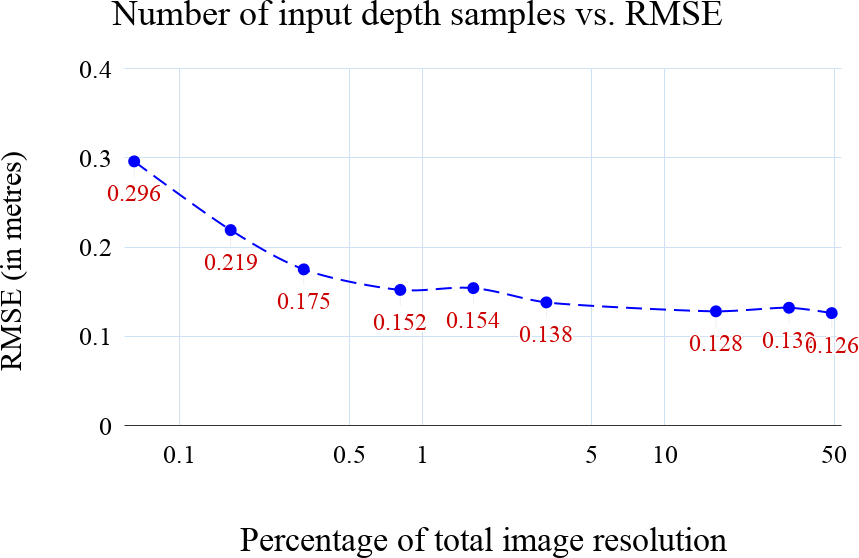}
\end{center}
   \caption{Number of input depth samples vs RMSE on NYUDepthV2 as a percentage of total image resolution. Note the use of log scale in the X-axis.}
\label{fig:samplesvserror}
\end{figure}

For this experiment, we use the NYUDepthV2 dataset as we are provided with dense ground truth information resulting in more consistent accuracy results. We train a different model for every sample size, limiting the training time to a fixed number of epochs each. We initialize our model with weights learned from our KITTI Depth Completion dataset to reduce our training time. We evaluate RMSE values on our validation set and a plot of this can be seen in Figure \ref{fig:samplesvserror}. As observed by Ma \etal in their network, the performance gained by adding more sparse input samples tends to saturate. We notice a saturation at around 5000 depth samples, roughly 1.7\% of the image resolution. Qualitatively we can see in Figure \ref{fig:nyudepth_images} that even with an extremely sparse input sample set, the RGB branch of our network is able to guide the depth prediction using mostly image based contextual cues.

\subsection{Penn Driving LiDAR RGB Dataset}

\begin{figure}[t]
\begin{center}
\includegraphics[width=110px]{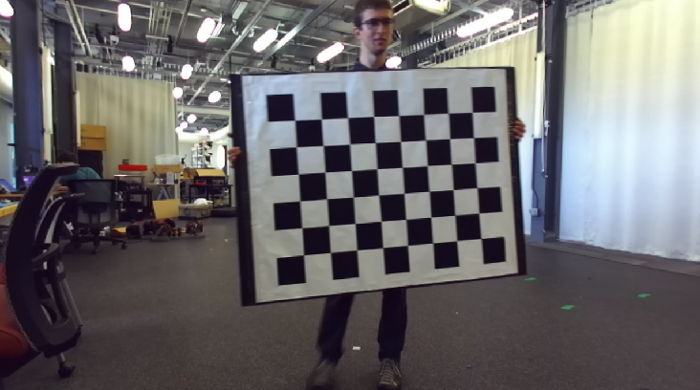}
\includegraphics[width=110px]{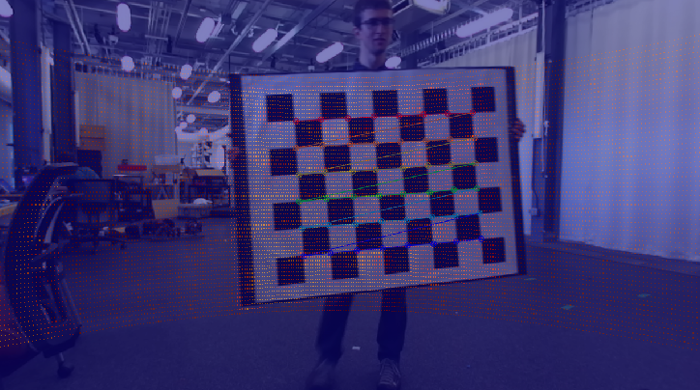}
\includegraphics[width=220px]{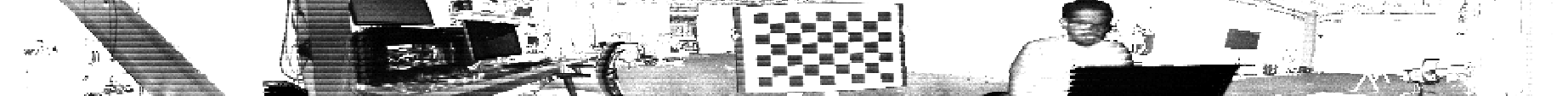}
\end{center}
   \caption{LiDAR RGB Calibration; Top: Rectified RGB image from the ZED Mini camera and projected LiDAR points onto the RGB Image after calibration. Bottom: Intensity image from Ouster OS-1 LiDAR from which we detect checkerboard corners.}
\label{fig:lidargbcalib}
\end{figure}

We additionally test this model on our own well-calibrated LiDAR and RGB dataset.  We use an Ouster OS-1\footnote{http://www.ouster.io} 64 beam LiDAR along with a StereoLabs Zed Mini\footnote{https://www.stereolabs.com/zed-mini/} camera.  In order to calibrate the extrinsics between the LiDAR and camera, we utilize the intensity image from the Ouster.  We detect checkerboard corners in the intensity image, and then project them into the point cloud to find their locations in 3 dimensional space.  These points can then be projected back onto a virtual image plane located in front of the LiDAR and standard stereo calibration tools can be used to calibrate the virtual LiDAR camera to the RGB camera.  We finally project the LiDAR point cloud onto the RGB image in order to verify the calibration, as shown in Figure \ref{fig:lidargbcalib}

Following calibration, we mounted the camera and LiDAR assembly to a car and collected 15 minutes of driving in downtown Philadelphia.  We artificially remove points from the point cloud to simulate a lower resolution sensor and run our pre-trained model. This translates to 5000 randomly sampled points from roughly 40,000 LiDAR points. The withheld points are then used for evaluating inference accuracy.

\section{Discussion and Conclusion}

In this section we discuss our observations and the motivation behind our design decisions in the context of datasets such as KITTI and NYU Depth.

\subsection{Architecture}

Jaritz \etal talk briefly about the benefits of a late fusion architecture over an early fusion one~\cite{jaritz2018sparse}. We agree with their statement and reaffirm the belief that given the different representations of RGB and depth modalities, the correct way to jointly combine this information is by learning to first transform it into a common feature space. While previous work has proposed single path architectures, where RGB and the sparse depth channels are concatenated into a single 4D input and passed to a network~\cite{mal2018sparse}, we propose the use of a number of individual and independent convolution and pyramid pooling operators on the individual modalities in a dual branch manner. We experimented with implementations where both modalities were fused prior to the SPP blocks and noticed a drop in performance, hinting that the additional independent information learned was useful to the final fusion and prediction. Figure \ref{fig:dual_branch} shows the information gained from having two branches in our network.

In terms of input sparsity, we experimented with replacing all our convolutions in the depth branch with sparse convolutions~\cite{uhrig2017sparsity} but noticed a significant drop in performance. Huang \etal propose the use of additional sparse operations such as sparsity invariant upsampling, addition and concatenation in addition to convolution and were able to achieve much better results~\cite{huang2018hms}. However, we are more inclined to believe that desirable performance can be achieved with the use of regular convolutions and operations for multi-modal input with simple pre-processing hole filling operations such as morphological filters, fill maps and nearest neighbor interpolation~\cite{dimitrievski2018learning, ku2018defense, chen2018estimating}. This is simple and effective in providing the network with a good initialization. 

We did notice however that with hole filling pre-processing steps, care must be taken in the use of residual connections from the depth channels to the penultimate layers. We found that using a residual connection from the second and third layers of our depth channel to the penultimate layer of our deconvolution layers led to similar accuracy as IPBasic~\cite{ku2018defense} but the network failed to learn to use information from the RGB branch. Perhaps such a network must be more carefully trained with carefully selected hyperparameters. However, in a single channel or early fusion network, adding residual connections from the depth input to the final layers has been shown to be highly effective~\cite{ma2018self,cheng2018depth}.
\subsection{Generalization}

\begin{figure}[t]
\begin{center}
\includegraphics[height=47px]{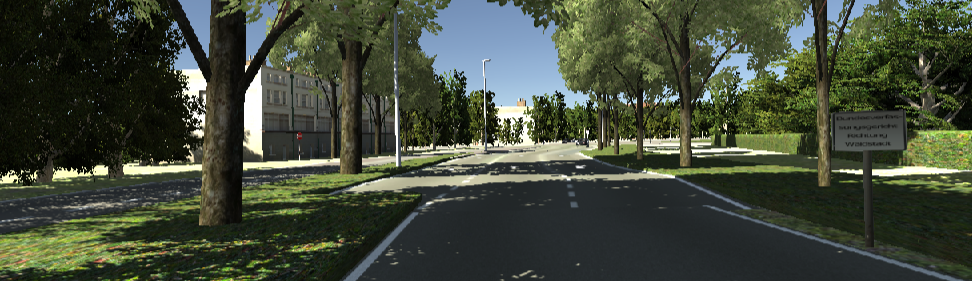}
\includegraphics[height=47px]{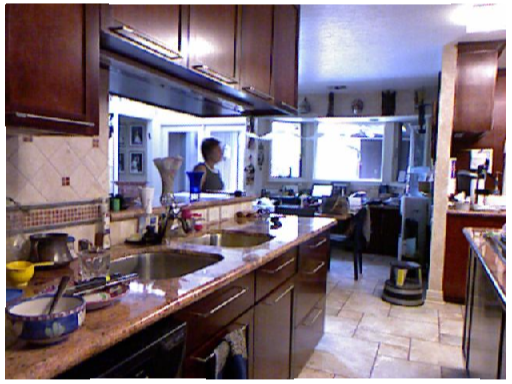}
\includegraphics[height=47px]{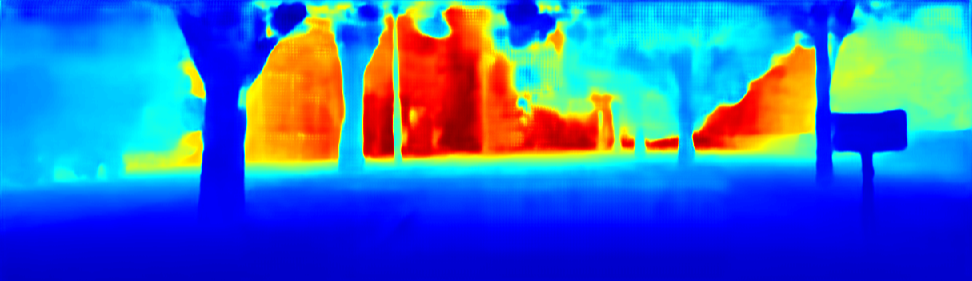}
\includegraphics[height=47px]{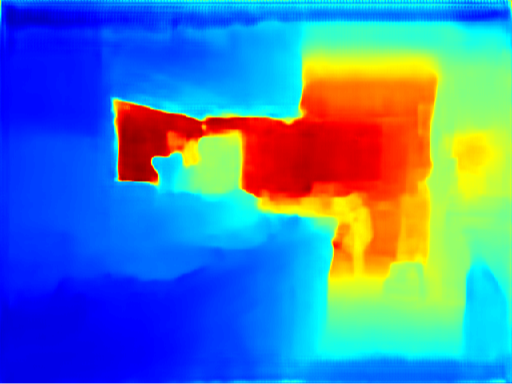}
\includegraphics[height=47px]{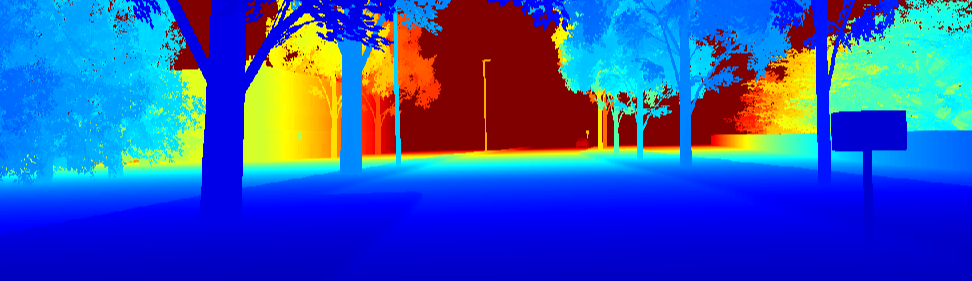}
\includegraphics[height=47px]{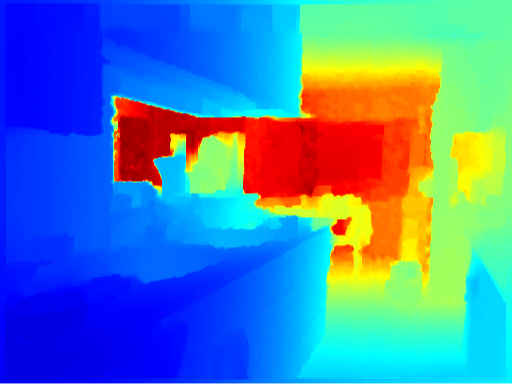}
\end{center}
   \caption{Generalization across datasets: Our model trained on the KITTI dataset is able generalize to new datasets such as Virtual KITTI (left) and NYUDepthV2 (right). No retraining or fine-tuning was performed. From top to bottom is the color image, predicted depth image and the ground truth.}
\label{fig:generalization}
\end{figure}

\begin{figure}[t]
\begin{center}
\includegraphics[width=110px]{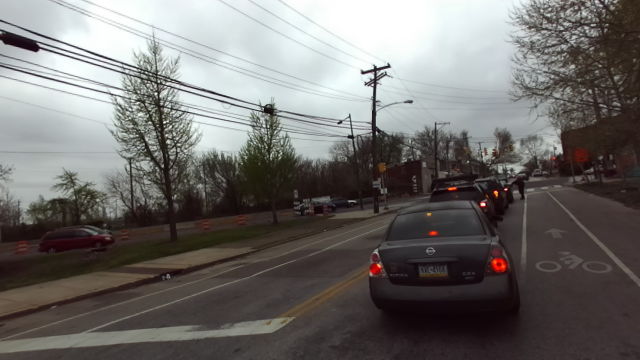}
\includegraphics[width=110px]{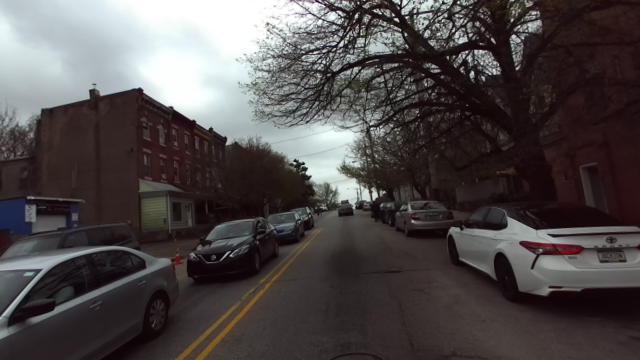}
\includegraphics[width=110px]{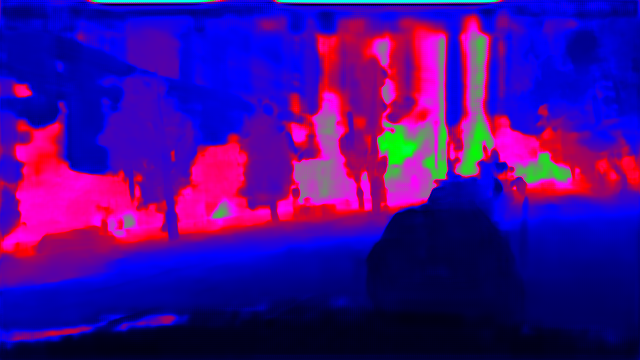}
\includegraphics[width=110px]{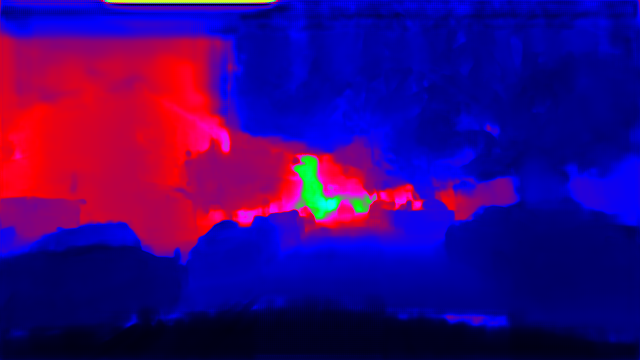}
\includegraphics[width=110px]{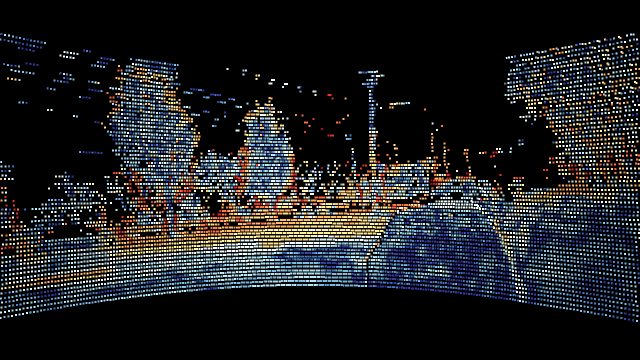}
\includegraphics[width=110px]{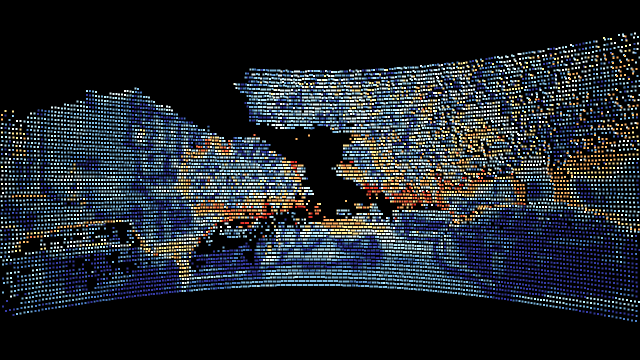}
\includegraphics[width=110px]{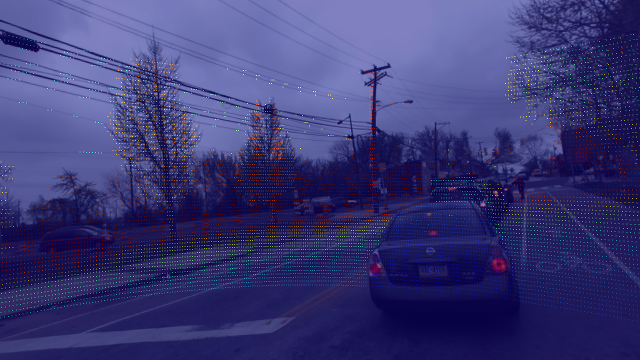}
\includegraphics[width=110px]{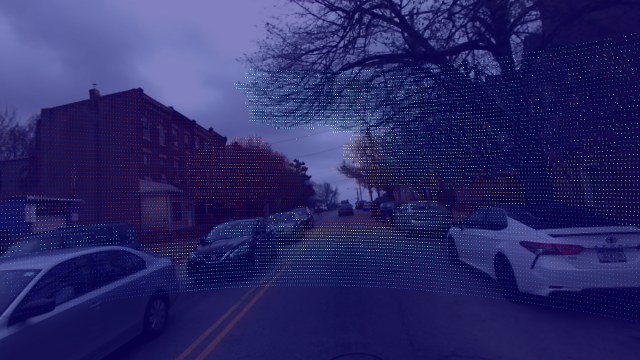}
\end{center}
   \caption{Penn Driving LiDAR RGB Dataset; Top-Bottom: Input RGB Image, predicted dense depth image, error image and overlay of LiDAR points on RGB for visualization. The predicted depth is generated using the model trained on KITTI data only. Colormaps for depth and error images are from the KITTI toolkit.}
\label{fig:penndrive}
\end{figure}

We also test the model we trained for the KITTI Depth Completion benchmark on Virtual KITTI, our Penn Driving dataset and NYUDepthV2. Figure \ref{fig:generalization} shows a few examples of predictions made using our KITTI Depth Completion model on Virtual Kitti and NYUDepthV2. Qualitatively, we noticed that the network is able to use sufficient RGB cues to generate semantically valid depth predictions but quantitatively we noticed errors in predicted depth values. This is due to the difference in depth scale across the three datasets (KITTI maximum depth is \SI{85}{m}, virtual kitti is \SI{655.3}{m} and NYUDepth is roughly \SI{10}{m}) and is quickly corrected after minimal training using the KITTI model as an initialization. We believe that the separate RGB image branch and the density of features learned independently in this branch to be a large contributor to the generalization of this network.  Our model trained on KITTI is able to achieve an RMSE value of \SI{0.607}{m} and MAE of \SI{0.216}{m} on our NYUDepthV2 test set using 10\% of ground truth as input depth samples. On the Penn driving dataset, we achieved an RMSE of \SI{6.690}{m} and MAE of \SI{3.157}{m}. The higher error is mainly in regions far away as shown in the error images in Figure \ref{fig:penndrive}.

\subsection{Conclusion}

We have proposed a CNN architecture that can be used to upsample sparse range data using the available high resolution intensity imagery. Our architecture is designed to extract contextual cues from the image to guide the upsampling process which leads to a network with separate branches for the image and depth data. We have demonstrated its performance on relevant datasets and shown that the approach appears to capture salient cues from the image data and produce upsampled depth results that respect relevant image boundaries and correlate well with the available ground truth. We also propose our own dataset for additional validation.

\balance 


\bibliographystyle{IEEEtran}
\bibliography{root}

\begin{thebibliography}{10}
\providecommand{\url}[1]{#1}
\csname url@rmstyle\endcsname
\providecommand{\newblock}{\relax}
\providecommand{\bibinfo}[2]{#2}
\providecommand\BIBentrySTDinterwordspacing{\spaceskip=0pt\relax}
\providecommand\BIBentryALTinterwordstretchfactor{4}
\providecommand\BIBentryALTinterwordspacing{\spaceskip=\fontdimen2\font plus
\BIBentryALTinterwordstretchfactor\fontdimen3\font minus
  \fontdimen4\font\relax}
\providecommand\BIBforeignlanguage[2]{{%
\expandafter\ifx\csname l@#1\endcsname\relax
\typeout{** WARNING: IEEEtran.bst: No hyphenation pattern has been}%
\typeout{** loaded for the language `#1'. Using the pattern for}%
\typeout{** the default language instead.}%
\else
\language=\csname l@#1\endcsname
\fi
#2}}

\bibitem{ferstl2013image}
D.~Ferstl, C.~Reinbacher, R.~Ranftl, M.~R{\"u}ther, and H.~Bischof, ``Image
  guided depth upsampling using anisotropic total generalized variation,'' in
  \emph{2013 IEEE International Conference on Computer Vision}.\hskip 1em plus
  0.5em minus 0.4em\relax IEEE, 2013, pp. 993--1000.

\bibitem{godard2017unsupervised}
C.~Godard, O.~Mac~Aodha, and G.~J. Brostow, ``Unsupervised monocular depth
  estimation with left-right consistency,'' in \emph{CVPR}, vol.~2, no.~6,
  2017, p.~7.

\bibitem{zhang2018deep}
Z.~Zhang, C.~Xu, J.~Yang, Y.~Tai, and L.~Chen, ``Deep hierarchical guidance and
  regularization learning for end-to-end depth estimation,'' \emph{Pattern
  Recognition}, vol.~83, pp. 430--442, 2018.

\bibitem{li2018monocular}
B.~Li, Y.~Dai, and M.~He, ``Monocular depth estimation with hierarchical fusion
  of dilated cnns and soft-weighted-sum inference,'' \emph{Pattern
  Recognition}, 2018.

\bibitem{li2018deep}
R.~Li, K.~Xian, C.~Shen, Z.~Cao, H.~Lu, and L.~Hang, ``Deep attention-based
  classification network for robust depth prediction,'' \emph{arXiv preprint
  arXiv:1807.03959}, 2018.

\bibitem{li2018megadepth}
Z.~Li and N.~Snavely, ``Megadepth: Learning single-view depth prediction from
  internet photos,'' in \emph{Computer Vision and Pattern Recognition (CVPR)},
  2018.

\bibitem{fu2018deep}
H.~Fu, M.~Gong, C.~Wang, K.~Batmanghelich, and D.~Tao, ``Deep ordinal
  regression network for monocular depth estimation,'' in \emph{IEEE Conference
  on Computer Vision and Pattern Recognition (CVPR)}, 2018.

\bibitem{song2018edgestereo}
X.~Song, X.~Zhao, H.~Hu, and L.~Fang, ``Edgestereo: A context integrated
  residual pyramid network for stereo matching,'' \emph{arXiv preprint
  arXiv:1803.05196}, 2018.

\bibitem{chang2018pyramid}
J.-R. Chang and Y.-S. Chen, ``Pyramid stereo matching network,'' in \emph{IEEE
  Conference on Computer Vision and Pattern Recognition}, 2018.

\bibitem{zbontar2016stereo}
J.~Zbontar and Y.~LeCun, ``Stereo matching by training a convolutional neural
  network to compare image patches,'' \emph{Journal of Machine Learning
  Research}, vol.~17, no. 1-32, p.~2, 2016.

\bibitem{kendall2017end}
A.~Kendall, H.~Martirosyan, S.~Dasgupta, and P.~Henry, ``End-to-end learning of
  geometry and context for deep stereo regression,'' in \emph{Computer Vision
  (ICCV), 2017 IEEE International Conference on}.\hskip 1em plus 0.5em minus
  0.4em\relax IEEE, 2017, pp. 66--75.

\bibitem{uhrig2017sparsity}
J.~Uhrig, N.~Schneider, L.~Schneider, U.~Franke, T.~Brox, and A.~Geiger,
  ``Sparsity invariant cnns,'' in \emph{2017 International Conference on 3D
  Vision (3DV)}.\hskip 1em plus 0.5em minus 0.4em\relax IEEE, 2017, pp. 11--20.

\bibitem{mal2018sparse}
F.~Ma and S.~Karaman, ``Sparse-to-dense: Depth prediction from sparse depth
  samples and a single image,'' in \emph{2018 IEEE International Conference on
  Robotics and Automation (ICRA)}.\hskip 1em plus 0.5em minus 0.4em\relax IEEE,
  2018, pp. 1--8.

\bibitem{ku2018defense}
J.~Ku, A.~Harakeh, and S.~L. Waslander, ``In defense of classical image
  processing: Fast depth completion on the cpu,'' \emph{arXiv preprint
  arXiv:1802.00036}, 2018.

\bibitem{eldesokey2018propagating}
A.~Eldesokey, M.~Felsberg, and F.~S. Khan, ``Propagating confidences through
  cnns for sparse data regression,'' \emph{arXiv preprint arXiv:1805.11913},
  2018.

\bibitem{eigen2014depth}
D.~Eigen, C.~Puhrsch, and R.~Fergus, ``Depth map prediction from a single image
  using a multi-scale deep network,'' in \emph{Advances in neural information
  processing systems}, 2014, pp. 2366--2374.

\bibitem{liu2016learning}
F.~Liu, C.~Shen, G.~Lin, and I.~D. Reid, ``Learning depth from single monocular
  images using deep convolutional neural fields.'' \emph{IEEE Trans. Pattern
  Anal. Mach. Intell.}, vol.~38, no.~10, pp. 2024--2039, 2016.

\bibitem{chen2016single}
W.~Chen, Z.~Fu, D.~Yang, and J.~Deng, ``Single-image depth perception in the
  wild,'' in \emph{Advances in Neural Information Processing Systems}, 2016,
  pp. 730--738.

\bibitem{kong2018pixel}
S.~Kong and C.~Fowlkes, ``Pixel-wise attentional gating for parsimonious pixel
  labeling,'' \emph{arXiv preprint arXiv:1805.01556}, 2018.

\bibitem{yang2018segstereo}
G.~Yang, H.~Zhao, J.~Shi, Z.~Deng, and J.~Jia, ``Segstereo: Exploiting semantic
  information for disparity estimation,'' in \emph{European Conference on
  Computer Vision}.\hskip 1em plus 0.5em minus 0.4em\relax Springer, 2018, pp.
  660--676.

\bibitem{khamis2018stereonet}
S.~Khamis, S.~Fanello, C.~Rhemann, A.~Kowdle, J.~Valentin, and S.~Izadi,
  ``Stereonet: Guided hierarchical refinement for real-time edge-aware depth
  prediction,'' in \emph{European Conference on Computer Vision}.\hskip 1em
  plus 0.5em minus 0.4em\relax Springer, 2018, pp. 596--613.

\bibitem{yang2014single}
C.-Y. Yang, C.~Ma, and M.-H. Yang, ``Single-image super-resolution: A
  benchmark,'' in \emph{European Conference on Computer Vision}.\hskip 1em plus
  0.5em minus 0.4em\relax Springer, 2014, pp. 372--386.

\bibitem{yamanaka2017fast}
J.~Yamanaka, S.~Kuwashima, and T.~Kurita, ``Fast and accurate image super
  resolution by deep cnn with skip connection and network in network,'' in
  \emph{Neural Information Processing}.\hskip 1em plus 0.5em minus 0.4em\relax
  Springer, 2017, pp. 217--225.

\bibitem{hui2016depth}
T.-W. Hui, C.~C. Loy, and X.~Tang, ``Depth map super-resolution by deep
  multi-scale guidance,'' in \emph{European Conference on Computer
  Vision}.\hskip 1em plus 0.5em minus 0.4em\relax Springer, 2016, pp. 353--369.

\bibitem{yeh2017semantic}
R.~A. Yeh, C.~Chen, T.-Y. Lim, A.~G. Schwing, M.~Hasegawa-Johnson, and M.~N.
  Do, ``Semantic image inpainting with deep generative models.'' in
  \emph{CVPR}, vol.~2, no.~3, 2017, p.~4.

\bibitem{yu2018generative}
J.~Yu, Z.~Lin, J.~Yang, X.~Shen, X.~Lu, and T.~S. Huang, ``Generative image
  inpainting with contextual attention,'' \emph{arXiv preprint}, 2018.

\bibitem{huang2018hms}
Z.~Huang, J.~Fan, S.~Yi, X.~Wang, and H.~Li, ``Hms-net: Hierarchical
  multi-scale sparsity-invariant network for sparse depth completion,''
  \emph{arXiv preprint arXiv:1808.08685}, 2018.

\bibitem{dimitrievski2018learning}
M.~Dimitrievski, P.~Veelaert, and W.~Philips, ``Learning morphological
  operators for depth completion,'' in \emph{International Conference on
  Advanced Concepts for Intelligent Vision Systems}.\hskip 1em plus 0.5em minus
  0.4em\relax Springer, 2018, pp. 450--461.

\bibitem{ronneberger2015u}
O.~Ronneberger, P.~Fischer, and T.~Brox, ``U-net: Convolutional networks for
  biomedical image segmentation,'' in \emph{International Conference on Medical
  image computing and computer-assisted intervention}.\hskip 1em plus 0.5em
  minus 0.4em\relax Springer, 2015, pp. 234--241.

\bibitem{van2019sparse}
W.~Van~Gansbeke, D.~Neven, B.~De~Brabandere, and L.~Van~Gool, ``Sparse and
  noisy lidar completion with rgb guidance and uncertainty,'' \emph{arXiv
  preprint arXiv:1902.05356}, 2019.

\bibitem{cheng2018depth}
X.~Cheng, P.~Wang, and R.~Yang, ``Depth estimation via affinity learned with
  convolutional spatial propagation network,'' in \emph{European Conference on
  Computer Vision}.\hskip 1em plus 0.5em minus 0.4em\relax Springer, 2018, pp.
  108--125.

\bibitem{wang2018multi}
B.~Wang, Y.~Feng, and H.~Liu, ``Multi-scale features fusion from sparse lidar
  data and single image for depth completion,'' \emph{Electronics Letters},
  2018.

\bibitem{ma2018self}
F.~Ma, G.~V. Cavalheiro, and S.~Karaman, ``Self-supervised sparse-to-dense:
  Self-supervised depth completion from lidar and monocular camera,''
  \emph{arXiv preprint arXiv:1807.00275}, 2018.

\bibitem{bromley1994signature}
J.~Bromley, I.~Guyon, Y.~LeCun, E.~S{\"a}ckinger, and R.~Shah, ``Signature
  verification using a" siamese" time delay neural network,'' in \emph{Advances
  in neural information processing systems}, 1994, pp. 737--744.

\bibitem{he2016deep}
K.~He, X.~Zhang, S.~Ren, and J.~Sun, ``Deep residual learning for image
  recognition,'' in \emph{Proceedings of the IEEE conference on computer vision
  and pattern recognition}, 2016, pp. 770--778.

\bibitem{odena2016deconvolution}
\BIBentryALTinterwordspacing
A.~Odena, V.~Dumoulin, and C.~Olah, ``Deconvolution and checkerboard
  artifacts,'' \emph{Distill}, 2016. [Online]. Available:
  \url{http://distill.pub/2016/deconv-checkerboard}
\BIBentrySTDinterwordspacing

\bibitem{hirschmuller2005accurate}
H.~Hirschmuller, ``Accurate and efficient stereo processing by semi-global
  matching and mutual information,'' in \emph{Computer Vision and Pattern
  Recognition, 2005. CVPR 2005. IEEE Computer Society Conference on},
  vol.~2.\hskip 1em plus 0.5em minus 0.4em\relax IEEE, 2005, pp. 807--814.

\bibitem{zhou2017unsupervised}
T.~Zhou, M.~Brown, N.~Snavely, and D.~G. Lowe, ``Unsupervised learning of depth
  and ego-motion from video,'' in \emph{CVPR}, vol.~2, no.~6, 2017, p.~7.

\bibitem{vijayanarasimhan2017sfm}
S.~Vijayanarasimhan, S.~Ricco, C.~Schmid, R.~Sukthankar, and K.~Fragkiadaki,
  ``Sfm-net: Learning of structure and motion from video,'' \emph{arXiv
  preprint arXiv:1704.07804}, 2017.

\bibitem{gaidon2016virtual}
A.~Gaidon, Q.~Wang, Y.~Cabon, and E.~Vig, ``Virtual worlds as proxy for
  multi-object tracking analysis,'' in \emph{Proceedings of the IEEE conference
  on computer vision and pattern recognition}, 2016, pp. 4340--4349.

\bibitem{silberman2012indoor}
N.~Silberman, D.~Hoiem, P.~Kohli, and R.~Fergus, ``Indoor segmentation and
  support inference from rgbd images,'' in \emph{European Conference on
  Computer Vision}.\hskip 1em plus 0.5em minus 0.4em\relax Springer, 2012, pp.
  746--760.

\bibitem{liao2017parse}
Y.~Liao, L.~Huang, Y.~Wang, S.~Kodagoda, Y.~Yu, and Y.~Liu, ``Parse geometry
  from a line: Monocular depth estimation with partial laser observation,'' in
  \emph{Robotics and Automation (ICRA), 2017 IEEE International Conference
  on}.\hskip 1em plus 0.5em minus 0.4em\relax IEEE, 2017, pp. 5059--5066.

\bibitem{jaritz2018sparse}
M.~Jaritz, R.~De~Charette, E.~Wirbel, X.~Perrotton, and F.~Nashashibi, ``Sparse
  and dense data with cnns: Depth completion and semantic segmentation,'' in
  \emph{2018 International Conference on 3D Vision (3DV)}.\hskip 1em plus 0.5em
  minus 0.4em\relax IEEE, 2018, pp. 52--60.

\bibitem{chen2018estimating}
Z.~Chen, V.~Badrinarayanan, G.~Drozdov, and A.~Rabinovich, ``Estimating depth
  from rgb and sparse sensing,'' \emph{arXiv preprint arXiv:1804.02771}, 2018.

\end{thebibliography}

\end{document}